\documentclass[letterpaper, 10 pt, conference]{ieeeconf}

\IEEEoverridecommandlockouts                              

\title{\LARGE \bf
TOP: Time Optimization Policy for Stable and Accurate Standing Manipulation with Humanoid Robots
}

\author{Zhenghan Chen$^{1^*}$, Haocheng Xu$^{1^*}$, Haodong Zhang$^{1}$, Liang Zhang$^3$, He Li$^{1}$, Dongqi Wang$^{2}$, Jiyu Yu$^{1}$, Yifei Yang$^{1}$ \\ Zhongxiang Zhou$^{1}$, Rong Xiong$^{1}$ %
\thanks{*The first two authors contributed equally. 1 The State Key Laboratory of Industrial Control and Technology of Zhejiang University. 2 Department of Engineering Mechanics of Zhejiang University. 3 Zhejiang Humanoid Robot Innovation Center.  Rong Xiong is the corresponding author. rxiong@zju.edu.cn.} %
}

\usepackage{array}     
\usepackage{booktabs}  
\usepackage{amsmath}
\usepackage{amssymb}
\usepackage{hyperref}
\usepackage{dsfont}
\usepackage{bm}
\usepackage{colortbl}  
\usepackage{xcolor}
\usepackage{stfloats}
\usepackage{subfigure} 
\usepackage{float} 
\usepackage{multirow} 
\usepackage{graphicx}
\usepackage{subcaption}
\usepackage{cleveref}
\usepackage{mathrsfs}
\usepackage{threeparttable}
\usepackage[
    backend=biber,
    sorting=none
]{biblatex}

\definecolor{mypink}{RGB}{191,0,95}  
\usepackage[letterpaper,top=60pt,bottom=43pt,left=48pt,right=48pt]{geometry}

\begin{document}

\maketitle
\pagestyle{empty}  
\thispagestyle{empty} 


\begin{abstract}
Humanoid robots have the potential capability to perform a diverse range of manipulation tasks, but this is based on a robust and precise standing controller. Existing methods are either ill-suited to precisely control high-dimensional upper-body joints, or difficult to ensure both robustness and accuracy, especially when upper-body motions are fast. This paper proposes a novel time optimization policy (TOP), to train a standing manipulation control model that ensures balance, precision, and time efficiency simultaneously, with the idea of adjusting the time trajectory of upper-body motions but not only strengthening the disturbance resistance of the lower-body. Our approach consists of three parts. Firstly, we utilize motion prior to represent upper-body motions to enhance the coordination ability between the upper and lower-body by training a variational autoencoder (VAE). Then we decouple the whole-body control into an upper-body PD controller for precision and a lower-body RL controller to enhance robust stability. Finally, we train TOP method in conjunction with the decoupled controller and VAE to reduce the balance burden resulting from fast upper-body motions that would destabilize the robot and exceed the capabilities of the lower-body RL policy. The effectiveness of the proposed approach is evaluated via both simulation and real world experiments, which demonstrate the superiority on standing manipulation tasks stably and accurately. The project page can be found at \href{https://anonymous.4open.science/w/top-258F/}{\textcolor{mypink}{https://anonymous.4open.science/w/top-258F/}}.

\end{abstract}

\section{INTRODUCTION}
Humanoid robots are the most potential embodied agents for the purpose of liberating human-level labors, as they are designed to perform anthropomorphic motions and various whole-body loco-manipulation tasks, including industrial parts assembly, home service, etc.\cite{gu2025humanoid}. Their anthropomorphism naturally makes them more suitable than other specific robots to interact with environments, objects and humans to complete various physical tasks. Although rapid growth has been achieved in the field of humanoid robots\cite{ben2025homie}, it remains a challenge to execute various intricate tasks while maintaining balance and precision simultaneously due to the intrinsic instability characteristic of humanoid robot. 

\begin{figure}[ht]
    \centering
    \setlength{\abovecaptionskip}{0.0cm}
    \setlength{\belowcaptionskip}{-0.2cm}
	\subfigure{
		\begin{minipage}[t]{0.48\textwidth}
			\centering
			\includegraphics[width=1\textwidth]{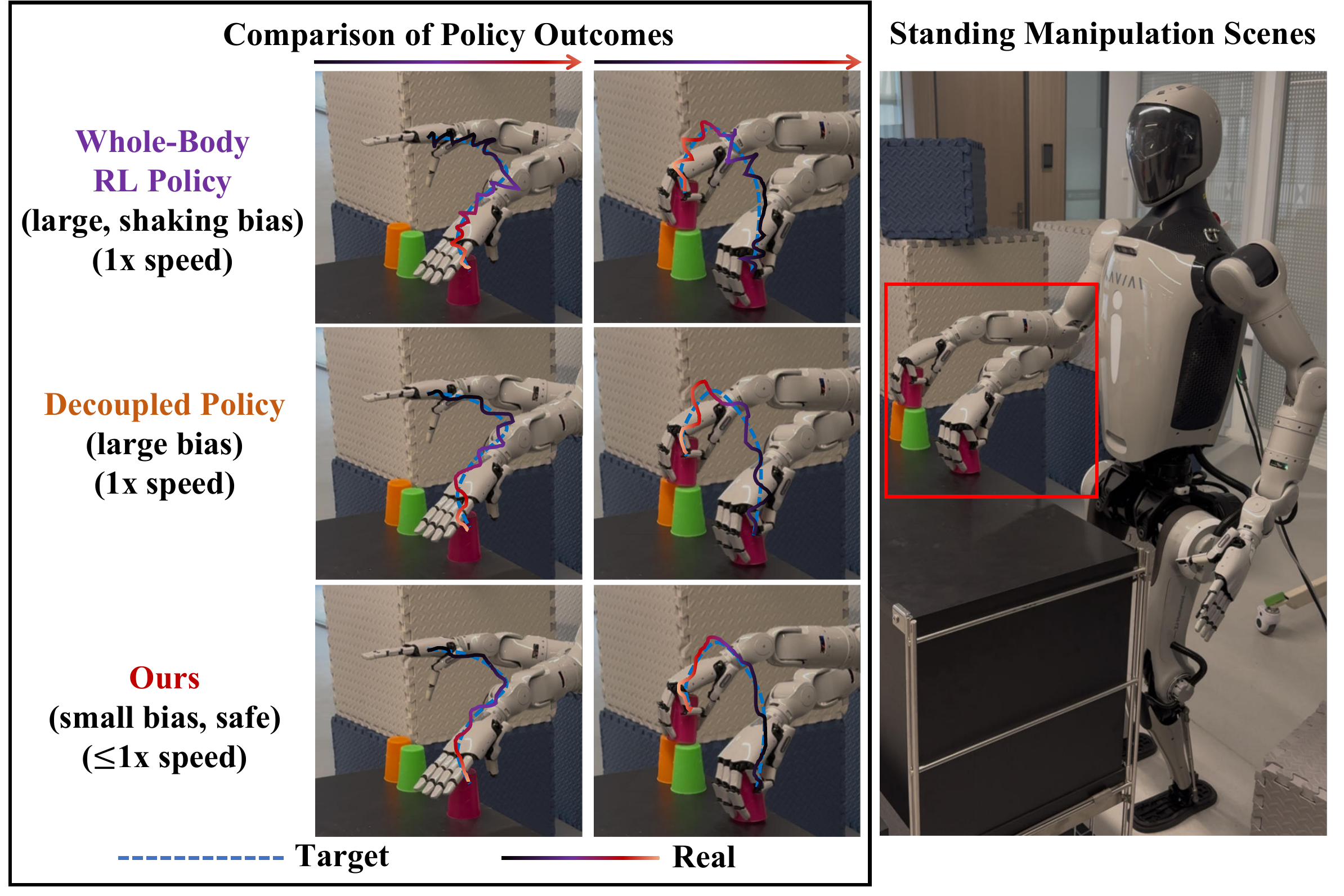}
		\end{minipage}
	}
    \caption{Illustration of different methods. \textbf{A}: Whole-body RL policy, but it will be ill-suited to control high-dimensional upper-body joint, which cause large and shaking bias from target trajectory. \textbf{B}: Decoupled policy avoid shaking bias by using PD controller for upper-body, but still lack the consideration about momentum caused by rapid upper-body motions, remaining large bias. \textbf{C}: Our method adjust timestamp of motions aiming to reduce the impact of momentum and making standing safer, which gain smaller bias but need more time to achieve the goal.}
    \label{fig::surface}
    \vspace{-15pt}
\end{figure}

Existing methods can be broadly divided into two paradigms: whole-body controllers\cite{romualdi2022online,serifi2024vmp,ji2024exbody2} and upper and lower-body decoupled controllers\cite{seo2023deep,lu2024mobile}. Traditional whole-body controllers, such as model predictive control (MPC), are capable of generating precise motions\cite{romualdi2022online}, but maintaining stability and robustness in real world remains a significant challenge. Whole-body RL-based controllers offer improved dynamic robustness\cite{radosavovic2024real,fu2024humanplus}, but often struggle to accurately track complex reference trajectories, particularly involving high-dimensional upper-body joints\cite{liu2024visual}, and are prone to overfitting to suboptimal behaviors or generating unpredictable actions. To mitigate these limitations, a decoupled control architecture has been proposed\cite{seo2023deep}, wherein the high-dimensional upper body is controlled by a PD controller to ensure precise trajectory tracking, while the lower body is governed by an RL policy to provide robust balance against external perturbations. This decoupled controller has shown promise in enhancing both the stability and precision required for standing manipulation tasks\cite{lu2024mobile}.

However, all the aforementioned methods tend to execute upper-body motions without fully considering the robot’s actual execution capabilities, as shown in \cref{fig::surface}. This oversight often neglects the dynamic consequences of momentum changes caused by fast upper-body motions, which can lead to instability, loss of balance, or even collisions with the environment. A key issue lies in the momentum introduced by upper-body motions: fast upper-body motions may destabilize the robot and impact the tracking precision, while slower motions can certainly reduce momentum changes and improve stability and accuracy, but sacrifice time efficiency. In other words, whether the reference trajectory is generated by teleoperation\cite{ben2025homie}, VLA\cite{liu2024rdt} or other planners\cite{zhao2023learning}, determining the appropriate motion speed remains a tricky problem in whole-body standing manipulation scenarios.

\begin{figure*}[ht] 
    \vspace{4pt}
    \centering
    \setlength{\abovecaptionskip}{0.2cm}
    \setlength{\belowcaptionskip}{-0.2cm}
    \includegraphics[width=0.95\textwidth]{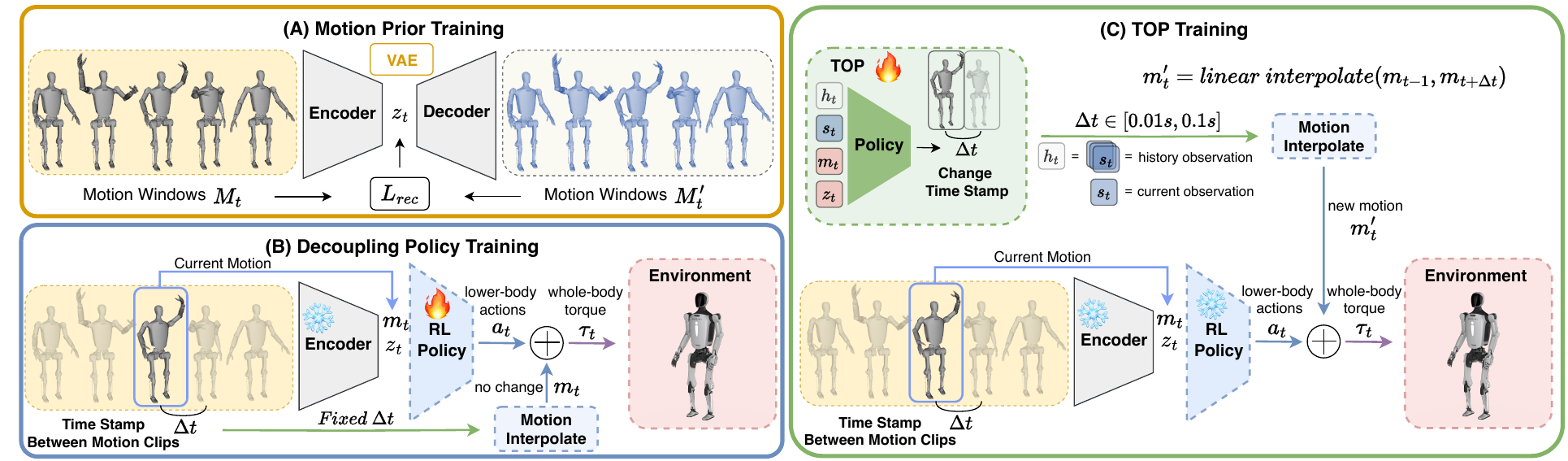}
    \caption{The overall architecture. (A) Training a latent code $z_t$ based on VAE structure to represent diverse upper-body motions. (B) A balance RL based policy to control the robot stay still, while upper-body use PD controller to execute current motion $m_t$, with a fixed time stamp $\Delta t$ between motions. (C) Then we train the TOP to optimize the timestamp between the motion clips to reduce the speed of motions and the impact of the momentum changes, which will slow down the current motion $m_t$ and execute new motion $m'_t$ calculated by $linear \ interpolate(m_{t-1}, m_{t+\Delta t})$.}
    \label{fig::framework}
    \vspace{-13pt}
\end{figure*}

Therefore, in this paper, we propose a novel learning-based framework to improve lower-body stability and upper-body motion precision by using \textbf{TOP} to optimize the timestamps of upper-body motions, minimizing momentum-induced balance disturbances. Our approach aims to strike a balance between stability, accuracy, and time efficiency by optimizing the time trajectory of motions to determine motion speed. This reduces the effect of fast upper-body motions on overall momentum and enhances both balance and execution efficiency, while minimizing disturbances from lower-body shifts that could affect upper-body tracking. The framework consists of three stages: first, a structured representation of diverse upper-body motions is learned to capture motion clip features; second, training a RL-based lower-body policy to maintain balance while the upper body executes motion clips via a PD controller; finally, training TOP method via supervised reinforcement learning to optimize the timestamps between motion clips. This alleviates balance challenges caused by fast upper-body motions that often exceed the limits of balance controllers, preventing falls and improving performance in standing manipulation tasks. We validate our approach in both simulation and real-world environments, demonstrating its versatility and effectiveness. Our contributions are as follows:

\begin{itemize}

\item propose a novel framework for humanoid standing manipulation that realizes stable, accurate, and time-efficient execution of complex upper-body motions and manipulation tasks.
\item As a key component of this framework, we introduce \textbf{TOP} (Time Optimization Policy), a supervised reinforcement learning module that optimizes motion timestamps to jointly enhance accuracy, stability, and efficiency.



\item our method is tested both in simulation and the real world that can generalize well to practical standing manipulation tasks, which demonstrates the adaptability and efficiency of our approach.



\end{itemize}

\section{RELATED WORKS}

\subsection{Whole-body Balance Controller}

The primary challenge of humanoid robots in achieving diverse rich and precision motions is the whole-body controller that tracks accurate upper-body motions and stabilizes the entire body in real time\cite{gu2025humanoid}. Previously, different attempts have been made in dynamics modeling and control\cite{ishiguro2017bipedal, penco2019multimode}, but these methods sacrifice precise models and fail to track large-scale motions. Moreover, the dilemma of balancing computationally efficiency and model complexity restricts its practical application\cite{romualdi2022online}.

Data-driven approaches are gradually taking the lead in this field, especially reinforcement learning combined with some sophisticated methods \cite{luo2023perpetual,hansen2024hierarchical}. Zhang et al. \cite{zhang2024whole} propose a method of whole-body locomotion by leveraging human motion references through imitation and optimization techniques, but it is difficult to generalize it to more whole-body motions. Fu et al. \cite{fu2024humanplus} introduce a framework based on transformer structures for humanoid robots to shadow and imitate human motions, improving natural interaction and adaptability. He et al. \cite{he2024omnih2o} successfully overcome the sim-to-real gap and train a universal and dexterous system for whole-body teleoperation and learning. However, these works focus on whole-body control with smaller arm DoFs and do not pay attention to the impact of lower-body stability on upper-body motion accuracy. 

To address the limitations of RL in precisely controlling high-dimensional upper-body joints, researchers adopt a decoupled control architecture that combines a PD controller for accurate upper-body motion execution with an RL-based lower-body controller to ensure robust stability. Cheng et al. \cite{cheng2024expressive} decouples the upper and lower-body controllers to obtain precision, but still lacks a measurement of the accuracy of upper-body motion and may lead to instability when the robot is standing still. Lu et al. \cite{lu2024mobile} propose a decoupled controller with CVAE to represent upper-body motions, aiming to maintain balance and precisely control in standing manipulation. On their basis, as a comparison, our approach integrates upper and lower-body separation control with time optimization policy that adjust the timestamp of upper-body motion, ensuring to improve standing stability, motion precision and time efficiency.

\subsection{Motion Representation Learning}
Motions from human or robots have spatial-temporal features in high-dimensional spaces\cite{starke2022deepphase}, and an effective representation to extract these complex spatial-temporal structures is highly needed\cite{peng2022ase}. Peng et al. \cite{peng2021amp} introduced adversarial training to learn motion priors, improving reinforcement learning policy efficiency and generalization. Hassan et al. \cite{hassan2023synthesizing} present a similar adversarial imitation learning framework to generate realistic interactions by integrating simulation with data-driven motion synthesis. However, when faced with large-scale data, it is difficult for such methods to learn a universal strategy, failed with mode collapse\cite{tang2024humanmimic}. 

Another common way to represent complex motion is to filter and extract key motion information to represent smooth temporal and spatial composition\cite{harvey2020robust}, but struggle with handling highly diverse or long-range motions. Recently, advances in generative models, like diffusion models\cite{zhou2024emdm} or variational autoencoder(VAE)\cite{won2022physics}, provided a good idea for representation of motions by creating a latent code to learn the distribution of multimodal motion sequences \cite{serifi2024vmp}. Tevet et al.\cite{tevet2022human} uses a transformer-based diffusion framework to generate natural human motions by directly predicting samples with geometric constraints. Li et al.\cite{li2025genmo} unifies human motion estimation and generation in a single framework, leveraging regression and diffusion to handle diverse conditions and tasks jointly. To provide a multimodal representation of upper-body motions, we decide to use VAE to extract a latent kinematic motion space and reconstruct them for training. 


\section{METHOD}
\label{sec::method}
\subsection{Overview} 

Our framework as depicted in \Cref{fig::framework}. The motion prior models the multimodal distribution of upper-body motions to improve coordination between the upper- and lower-body controllers. In the decoupling controller training stage, we adopt a curriculum schedule to ease RL exploration and fix the timestamps of motion clips, using the motion prior to train a robust lower-body controller. In the TOP training stage, we introduce supervised reinforcement learning to optimize the time stamps, with initial guesses provided by supervised learning. Action chunking \cite{zhao2023learning} is employed to further smooth the time stamps, as detailed in \ref{sec::method_top}.

\subsection{Extract Motion Priors}
\label{sec::method_motion_priors}
To improve awareness of the lower-body controller for upper-body past and future motions, we provide prior knowledge of upper-body motion, which is important for training more robust lower-body controllers \cite{liu2024visual,cheng2024expressive}. Specifically, we train an encoder-decoder pair to reconstruct the sequence of upper-body motions and incorporate the latent space as a representation in the state space of the lower-body control.

To represent diversity kinematic motions of a human or robot and capture more fine-grained level of motions, we use variational autoencoder (VAE) structure, which has already been shown as an effective motion representation and has the ability to learn the motion distribution and similarities of adjacent motions. We extract kinematic state of upper-body motions, consisting of joint position, velocity for a few past and future window frames. Our VAE structures include an encoder $E$ and a decoder $D$, and the latent space is modeled as a multivariate Gaussian distribution $\bm{z}_t \in \mathbb{R}^{d_z}$. 

\begin{equation}
\label{Eq:motion_clip}
    \bm{m}_t = \{ \bm{r}_t, \bm{\theta}_t, \bm{q}^{upper}_t, \bm{\dot{q}}^{upper}_t\}.
\end{equation}

\noindent where $\bm{r}_t \in \mathbb{R}^3$ is the position of the base relative to the world frame. Since we only need to encode the upper-body motion, setting the $\bm{r_t}$ as the constant is natural. $\bm{\theta}_t$ is the orientation of the base frame, represents as 6D vector. The joint angles and velocities are given by $\bm{q}^{upper}_t \in \mathbb{R}^{n_j}$ and $\bm{\dot{q}}^{upper}_t \in \mathbb{R}^{n_j}$, where $n_j = 15$ includes two 7-dof arms and one waist joint.

Formally, we extract the past and future frames to consist motion windows of length $2W+1$ from the distinct motion clips. In order to ensure the stability of the training of VAE networks, it is common to normalize the input data, and we use the mean and standard deviation of all motion clips in dataset to normalize our motion window frames, except orientation.

\begin{equation}
\label{Eq:motion_window}
    \bm{M}_t = \{ \bm{m}_{t-W}, \dots,\bm{m}_t, \dots, \bm{m}_{t+W} \}.
\end{equation}

The encoder of our VAE $E_{\phi}(\bm{z}_t|\bm{M}_t)$ maps the motion window to latent space $\bm{z}_t \in \mathbb{R}^{d_z}$, $d_z = 64$, and the sampled latent variable is then mapped back to input space by the decoder $D_{\theta}(\bm{M}'_t|\bm{z}_t)$. And we decide to use $\beta$-VAE\cite{higgins2017beta} with the reconstruction loss as follows:
\begin{align}
    L_{rec}(\bm{M}_t,\bm{M}'_t) &= \frac{1}{2W+1} \sum_{i=t-W}^{t+W}l_{rec}(\bm{m}_t,\bm{m}'_t) \\
    l_{rec}(\bm{m}_t,\bm{m}'_t) &= ||R(\bm{\theta}_t)-R(\bm{\theta}'_t)|| + ||\bm{q}_t-\bm{q}'_t||  \nonumber \\
    &\quad + ||\bm{\dot{q}}_t-\bm{\dot{q}}'_t|| + ||\bm{p}_t-\bm{p}'_t||
\end{align}

\noindent where the $R(\cdot)$ represents that computing rotation matrices for orientations using the Gram-Schmidt process, and because we already normalize the quantities, no relative weights are needed here. It is worth noting that the window size of $\bm{M}_t$ should be short enough to achieve motion generalization that the latent space can capture the features of primitive motion blocks that may appear in unseen motion sequences.

\subsection{Training Decoupling Policy}
\label{sec::method_rl}
We train a balance policy using a Legged Gym-based reinforcement learning framework, where PPO was used to update our lower-body policy. During training, the motion sequence is randomly chosen from the dataset at the beginning of a new episode, and retrieving the motion pair $(\bm{m}_t, \bm{z}_t)$ from Encoder $E_{\phi}(\bm{z}_t|\bm{M}_t)$. Then, we feed the motion pair to our lower-body policy to provide the instantaneous kinematic reference motions of upper-body and the past and future information, which helps the comprehension of the policy about disturbances caused by upper motions on balance. 

We consider our balance lower control policy as a goal-conditional $\pi_{\phi}(\bm{a}_t|\bm{s}_t,\bm{g}_t):\mathbb{G} \times \mathbb{S} \rightarrow \mathbb{A}$, where $\bm{g}_t \triangleq (\bm{m}_t, \bm{z}_t)\in \mathbb{G}$ is the goal at the time $t$ that indicates the target of upper-body motion clip and the latent code from Encoder $E_{\phi}(z_t|M_t)$ from dataset. $\bm{s}_t \triangleq \{\bm{q}_t, \bm{\dot{q}}_t, \bm{\theta}_{t},\bm{\omega}_t,\bm{a}_{t-1},\bm{g}_t\}\in \mathbb{S}$ is the current observation, where $\bm{q}_t \in \mathbb{R}^{27}$, $\bm{\dot{q}}_t \in \mathbb{R}^{27}$ are the position and velocity of whole body joints, $\bm{\omega}_t$ is the angular velocity of the base, $\bm{a}_{t-1}\in \mathbb{R}^{12}$ means the last action of lower-body joint. $\bm{a}_t \in \mathbb{A}$ is the action of lower-body joints. Both of upper-body motions and lower-body actions are actuated by a PD torque controller $\bm{\tau}_t = k_p(\bm{a}_t-\bm{q}_{t}) + k_d\bm{\dot{q}}_t$ for each joint. The reward design is shown in \Cref{table_param}. It should be noted that the rewards for regularization of actions are used to shape the standing mode, and the input history will be encoded as a hybrid internal embedding\cite{long2023hybrid}, which improves the training efficiency and robustness of lower-body RL controller. 


In order to reduce the exploration burden caused by upper-body motions and gain more stable training process, we introduce a training curriculum schedule that will change the amplitude of target motion clips\cite{lu2024mobile}. For the PD controller of the joint position during the training, the target joint position is calculated by 
\begin{equation}
\label{Eq:curriculum}
    \bm{q}^{upper}_{t} = \bm{q}^{upper}_{default} + \alpha_i(\bm{q}^{upper}_{target} - \bm{q}^{upper}_{default})
\end{equation}
where $\bm{q}^{upper}_{default}$ is the default joint position of upper-body. The $\alpha_i \in [0,1]$ is the unique amplitude factors of motion $i$, which is changed during the training by the rules similar to the \cite{lu2024mobile}.


\subsection{Time Optimization Policy}
\label{sec::method_top}

We design a reinforcement learning policy to optimize the timestamp between motion clips $\bm{m}_t,\bm{m}_{t+1}...\bm{m}_{t+N}$, considering the latent variable $\bm{z}_t$, the current observation $\bm{s}_t$ and the history observation $\bm{h}_t$. The structure of this net can be described as $\Delta t^{seq}_t = \pi_{\theta}(\bm{m}_t, \bm{z}_t, \bm{s}_t, \bm{h}_t)$, the $\theta$ is the learnable variable of TOP. The output of TOP ($\Delta t^{seq}_t = \Delta t_t, ..., \Delta t_{t+N}$) will feed into RL policy combined with the motion pair $(\bm{m}_t, \bm{z}_t)$. Once the timestamp of motion has been set to ${t+\Delta t_t}$, it is necessary to change the origin motion $\bm{m}_t$ to the new motion $\bm{m}'_t = linear \ interpolate(\bm{m}_{t-1},\bm{m}_{t+\Delta t_t})$ by linear interpolation. Because the motions of the dataset $\mathcal{M}$ satisfy the kinematic and dynamic constraints of the robot, after linear interpolation, $m'_t$ will not violate the constraints.

It should be noted that because of the rapid motion in the past few frames, the robot may lose its balance at the current moment, which exhibits a certain degree of lag. In other words, when we change the timestamp of the current upper-body motion to slow down the motion, it may be reflected in future multi frames feedback. This means that the output of the past policy ($\Delta t_{t-N}$) will affect the current balance performance of robot. Therefore, we are seeking to optimize timestamps for a period of time in the future, and inspired by \textit{action chunking}\cite{lai2022action}, the policy model becomes $\pi_{\theta}(\Delta t^{seq}_t|\bm{m}_t,\bm{z}_t,\bm{s}_t,\bm{h}_t)$ instead of $\pi_{\theta}(\Delta t_t|\bm{m}_t,\bm{z}_t,\bm{s}_t,\bm{h}_t)$, $N=10$ is the horizon step, which was obtained based on our testing.

A simple implementation of predicting a horizon of future timestamps like action chunking will be sub-optimal: if we directly shift by one timestamp until the end of trajectory, a new timestamps trajectory is incorporated abruptly every N steps and can result in the sudden action to slow down. To improve the action smoothness and avoid jerky discrete switching of timestamps, we query the output of policy at every timestamp, and give every prediction with an exponential weighting scheme $w_i = exp( -k * i)$, where $w_0$ represents the weight of the oldest action. Then we use the weighted average for the current predictions similar to \cite{zhao2023learning}.

Obviously, this kind of feedback lag brings an extra burden of policy exploration and it is more difficult to predict a short horizon of timestamp trajectory than a single timestamp. It is natural for us to think that the original timestamp of the motion clips can be set as the initial guess solution for the policy. Therefore, we introduce the supervise learning combined with reinforcement learning to reduce the need for inefficient random exploration, which also help steer the agent toward high-quality policies instead of suboptimal solutions \cite{wang2018supervised}. Our PPO-based supervised reinforcement learning framework incorporates the following loss components:
\begin{align}
    \label{eq::ppo_loss}
    &L_{sup} = -\sum\nolimits_i \bm{\Delta}t_i^{log}\bm{\pi}_{\theta}(\bm{\Delta}t_i^*|\bm{s}_i) \\
    &L_{RL} = -\mathbb{E}_t[\min(r_t(\theta)A_t, clip(r_t(\theta),1-\epsilon,1+\epsilon)A_t)] \\
    &L_{total} = \lambda_{sup}L_{sup} + \lambda_{RL}L_{RL}
\end{align}
Where $L_{sup},L_{RL}$ are the loss of supervise and reinforcement learning. $r_t(\theta)$ is the policy ratio, and $A_t$ is the advantage function. We combine both losses with weighting factors $\lambda_{sup} = 0.1$ and $\lambda_{RL} = 0.5$ to balance supervised learning and reinforcement learning. At the beginning of training, $L_{sup}$ is relatively large and plays a dominant role. After $L_{sup}$ decreases, the loss of $L_{RL}$ plays a dominant role. And the rewards of TOP is shown in \Cref{table_TOP}. 

We can easily observe that the reward design of TOP and the design of RL Rewards have similarities, as they both want to guide that robots can execute motions accurately while maintaining balance as much as possible. So why not train jointly? Because we want to make the TOP policy independent of the controller, we can retrain the top policy when the controller changes. In this way, we can freely change and update the controller without considering the top policy, and the performance of TOP policy will not impact the controller itself. During our training, we find that training both policy jointly will not affect the results of RL policy, but need more iteration to converge and the worse performance of TOP policy, which means separate training will bring more stable training results. 

\begin{table}[h]
    \setlength{\abovecaptionskip}{-0.0cm}
    \setlength{\belowcaptionskip}{-0.2cm}
    \caption{REWARDS OF TOP}
    \label{table_TOP}
    \centering
    \renewcommand{\arraystretch}{1.0} 
    \resizebox{0.47\textwidth}{!}{
    \begin{tabular}{ l l r }
        \toprule
        \textbf{Term} & \textbf{Expression} & \textbf{Weight} \\
        \midrule
        Gravity projection & $\exp(-20\| \mathbf{pg}_t^{xy} \|)$ & 2.5 \\
        Balance penalty & $\exp(20\sqrt{(\mathbf{pg}_t^x)^2+(\mathbf{pg}_t^y)^2})-1 $ & -1.0 \\
        Support constraint & $\lg(7*(\mathbf{p}^{center}_{feet}-\mathbf{p}^{project}_{com}))$ & -5 \\
        Encourage small $\Delta t$ & $\sum_{i=0}^{N}\exp\left(-\frac{(\Delta t_i)^2}{2\sigma^2}\right), \sigma=0.5$ & 5.0\\
        $\Delta t$ smooth & $ \sum_{i=0}^{N-1}(\| \Delta t_{i+1} - \Delta t_{i} \|)$ & -0.1 \\
        $\Delta t$ norm & $\| \sum_{i=0}^{N} \Delta t_i \|$ & 0.1 \\
        \bottomrule
    \end{tabular}
}
\vspace{-10pt}
\end{table}

\section{EXPERIMENTS}

\subsection{Experimental Setup}

We validate our approach on a full-size humanoid robot in both simulation and real-world. The robot is 1.65m tall, weighs 60kg, and has 41 DoFs, including two 7-DoF arms (6kg each) with a 3kg payload per arm. This combination of weight, precision, and manipulation capability makes balance during manipulation particularly challenging. The training dataset $\mathcal{M}$ is the large-scale GRAB human motion dataset \cite{taheri2020grab}. For testing, we additionally introduce more complex motions comprising 16,000 clips, forming dataset $\mathcal{T}$. All these motions are retargeted to humanoid robot via motion retargeting methods\cite{zhang2022kinematic}.

\noindent \textbf{VAE} We use a latent dimension of $d_z = 64$ and a window size of $W = 30$ (0.6s in real time). The VAE consists of four 1D convolutional layers with Layer Normalization, ReLU activations, and a final linear layer. At the bottleneck, the encoder output is doubled and sampled from a multivariate Gaussian \cite{serifi2024vmp}. We adopt a $\beta$-VAE objective with a KL weight of 0.002 and a cyclical schedule to prevent KL collapse, combined with cosine annealing with warm restarts of learning rate for stable convergence. See \Cref{table_param} for hyperparameter details.

\noindent \textbf{RL Policy} We apply domain randomization for robustness to object properties, similar to \cite{radosavovic2024real}. Base CoM position, motor delays, and torque noise are critical for addressing the sim-to-real gap. We also add upper-body position and velocity noise during training to improve the robustness of the lower-body balance controller.

\noindent \textbf{TOP} Our TOP network uses a three-layer MLP and an actor-critic framework for stable learning. Supervised learning provides an initial guess for $\Delta t^{seq}t$ to encourage smaller $\Delta t_t$ values. If the robot falls, the supervision loss $L{sup}$ is omitted from $L_{total}$, allowing the robot to slow down motions to maintain success. In practice, we limit $\Delta t$ to $[0.01s, 0.1s]$. We find that action chunking with a weighting scheme and the motion prior are both critical for producing smooth, precise motions. Results are shown in \cref{tab:simulation_results}.

\begin{table}[h]
    \setlength{\abovecaptionskip}{-0.0cm}
    \setlength{\belowcaptionskip}{-0.2cm}
    \caption{VAE PARAMETERS AND RL REWARDS}
    \label{table_param}
    \centering
    \renewcommand{\arraystretch}{1.0} 
    \resizebox{0.47\textwidth}{!}{
    \begin{tabular}{ l c | l r }
        \toprule
        \multicolumn{2}{c|}{\textbf{VAE Parameters}} & \multicolumn{2}{c}{\textbf{VAE Training}} \\
        \midrule
        \textbf{Param.} & \textbf{Value} & \textbf{Param.} & \textbf{Value} \\
        \midrule
        $kl \ weight$ & 0.002 & Batch size & 512 \\
        $W$ & 30 & Number of epochs & 30 000 \\
        $d_z$ & 64 & Learning rate & 0.003 \\
        Param. & 0.8M & KL-scheduler cycles/ratio & 7 / 0.5 \\
    \end{tabular}
    }
    \resizebox{0.47\textwidth}{!}{
    \begin{tabular}{ l l r }
        \toprule

        \textbf{Term} & \textbf{Expression} & \textbf{Weight} \\
        \midrule
        Base linear Velocity$^{xy}$ & $\exp{-4(\mathbf{v}^2_{xy})}$ & 3.0\\
        Base linear Velocity$^{z}$ & $\Sigma \mathbf{v}^2_{z}$ & -0.8\\
        Base angular Velocity$^{xy}$ & $\Sigma \bm{\omega}^2_{xy}$ & -0.1\\
        Base orientation & $\Sigma \mathbf{g}^2_{xy}$ & -1.5 \\
        Base acceleration & $\exp (-3 \| \mathbf{v}_t - \mathbf{v}_{t-1} \|_2 )$ & 0.2 \\
        Stand still & $\exp (-10 \| \mathbf{q}_t^{leg} - \mathbf{q}^{leg,ref} \|_2 )$ & 1.0 \\
        Feet contact & $\mathbf{1} ( F_{\text{feet}}^z \geq 5 )$ & 0.5 \\
        Feet slip & $\mathbf{1} ( F_{\text{feet}}^z \geq 5 ) \times \sqrt{\| \mathbf{v}_t^{\text{feet}} \|_2 }$ & 0.2 \\
        Action rate & $\| \mathbf{a}_t - \mathbf{a}_{t-1} \|_2^2$ & -0.2 \\
        Action acceleration & $\| \mathbf{a}_t + \mathbf{a}_{t-2} - 2\mathbf{a}_{t-1} \|_2^2$ & -0.2 \\
        Torques & $\| \boldsymbol{\tau}_t \|_2^2$ & -5e-6 \\
        DoF velocity & $\| \dot{\mathbf{q}}_t \|_2^2$ & -5e-4 \\
        DoF acceleration & $\| \ddot{\mathbf{q}}_t \|_2^2$ & -1e-7 \\
        \bottomrule
    \end{tabular}
    }
    \vspace{-10pt}
\end{table}

\subsection{Evaluation of Motion Priors and Decoupling Policy}
\begin{figure}[htbp]
      \vspace{5pt}
      \setlength{\abovecaptionskip}{-0.1cm}
      \setlength{\belowcaptionskip}{-0.2cm}
    \centering
	\subfigure{
        \rotatebox{90}{\scriptsize{~~~Motion2~~~~~\textbf{50\%Interp.}~~~~~Motion1}}
		\begin{minipage}[t]{0.45\textwidth}
			\centering
			\includegraphics[width=1\textwidth]{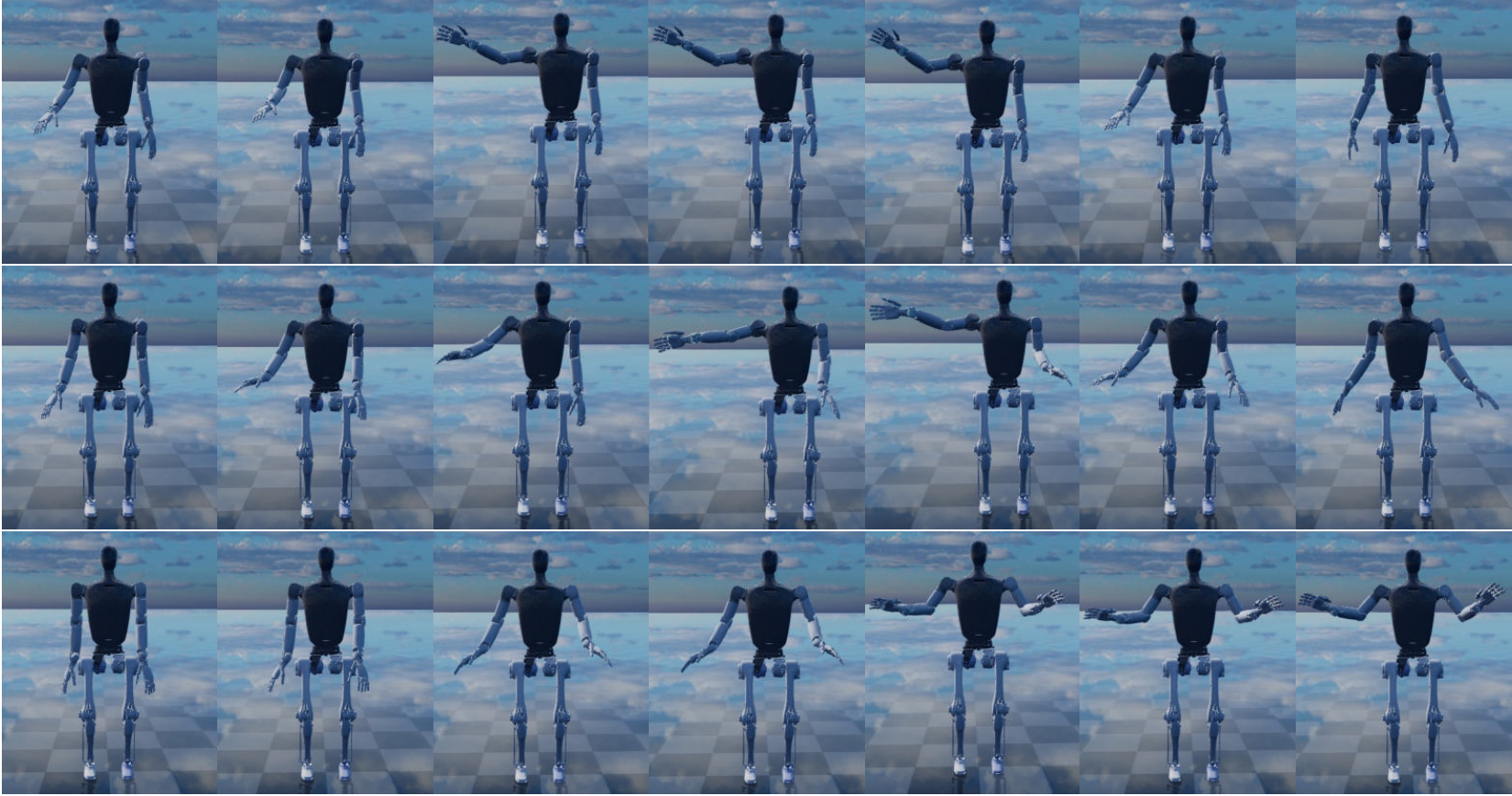}  
		\end{minipage}
	}
    \caption{\textbf{Latent Generalization.} Given two motion windows, one from the dataset $\mathcal{M}$ (row 1), the other from unseen dataset $\mathcal{T}$ (row 3), and the results of 50\% linear interpolation show in (row 2).}
    \label{fig::VAE_interp}
    \vspace{-10pt}
\end{figure}

\noindent\textbf{Generalization of Motion Priors.} 
To evaluate the efficacy of the VAE model and the latent space, we use linear interpolating between the latent space trajectories corresponding to two different motions, one motion comes from dateset $\mathcal{M}$, the other is an unseen motion from dataset $\mathcal{T}$. As shown in \Cref{fig::VAE_interp}, the result of 50\% linear interpolation between two different latent code, and we reconstruct it by Decoder $D_{\theta}(\bm{M}'_{t}|\bm{z}_t)$. We also show the reconstructed results of motions on the website, which demonstrate that our latent space can capture the features in short-horizon motions and has the ability to represent and reconstruct unseen motions that are in proximity to those of similar motion windows within the dataset.


\begin{figure}[htbp]
      \vspace{-5pt}
      \setlength{\abovecaptionskip}{-0.1cm}
      \setlength{\belowcaptionskip}{-0.0cm}
        \centering
	\subfigure{
        \rotatebox{90}{\scriptsize{~~~Conduct Slow~~~~~~~~~~~Pick}}
		\begin{minipage}[t]{0.45\textwidth}
			\centering
			\includegraphics[width=1\textwidth]{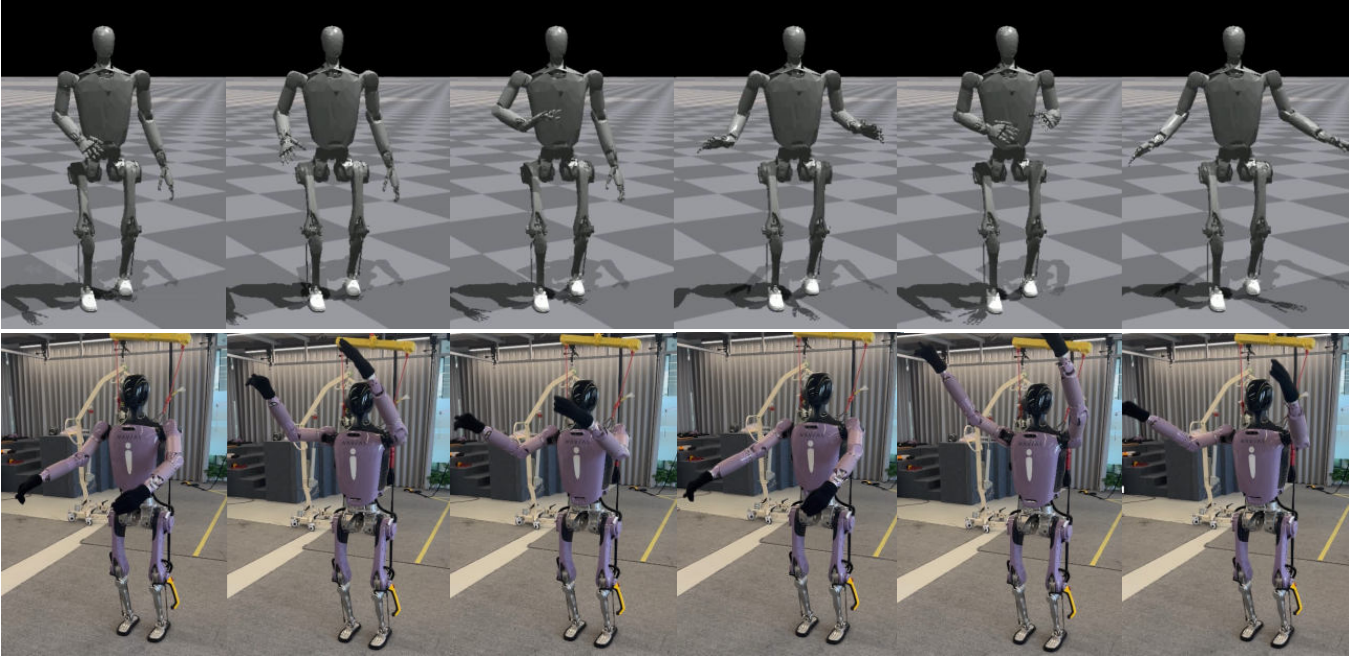}
		\end{minipage}
	}
    \caption{\textbf{Robustness.} Given motion windows in simulation and real world, pick motions are sampled from the dataset $\mathcal{M}$ (row 1), and the conduct slower motions comes from unseen dataset $\mathcal{T}$(row 2).}
    \label{fig::RL_evaluate}
    \vspace{-10pt}
\end{figure}

\noindent \textbf{Robustness of Decoupling policy.} To thoroughly evaluate the performance of our lower-body balance policy without TOP, we test upper-body motions from dataset $\mathcal{M}$ and unseen dataset $\mathcal{T}$. It is worth noting that the amplitude and speed of the motions from dataset $\mathcal{T}$ are larger than those in dataset $\mathcal{M}$, which put higher demands on the robustness. As shown in \Cref{fig::RL_evaluate}, our policy can maintain balance while tracking upper-body motions with high precision and balance RL policy demonstrates strong robustness. Although there are occasional failures on the unseen dataset $\mathcal{T}$, particularly in challenging cases where the motion is too fast or both arms are raised above the head resulting in significant momentum changes that destabilize the robot. By slowing down such difficult motions, our system achieves a success rate of over 80\%, indicating good generalization to unseen scenarios. The quantitative data are available in \Cref{tab:simulation_results}.


\subsection{Evaluation of TOP.}

\begin{table*}[t]
    \vspace{0pt}
    \centering
    \renewcommand{\arraystretch}{1.0}  
    \setlength{\tabcolsep}{5pt}  
    \resizebox{0.98\textwidth}{!}{
    \begin{tabular}{l|cc|ccc|ccc|cc}
        \toprule
        \textbf{Method} & 
        Time Cost $\downarrow$ & Success Rate $\uparrow$&
        $\mathbf{E^{\text{upper}}_{\text{jpe}}} \downarrow$ & 
        $\mathbf{E^{\text{upper}}_{\text{eepe}}} \downarrow$ & 
        $\mathbf{E^{\text{upper}}_{\text{eeoe}}} \downarrow$ & 
        $\mathbf{E_g} \downarrow$ & 
        $\mathbf{E^{\text{lower}}_{\text{acc}}} \downarrow$ & 
        $\mathbf{E^{\text{lower}}_{\text{action}}} \downarrow$ & 
        $\mathbf{E^{\text{upper}}_{\text{acc}}} \downarrow$ & 
        $\mathbf{E^{\text{upper}}_{\text{action}}} \downarrow$ \\
        \midrule
        \rowcolor{gray!30} \multicolumn{11}{l}{\textbf{Comparative Results}}\\
        \midrule
        \rowcolor{blue!30}
        Fixed Root (reference) & $15.0s$ & 100.0\% & 0.0130 & 0.0164 & 0.0594 &1.000 & - & - & - & -\\
        Exbody\cite{cheng2024expressive} & $\bm{15.0s}$ & 92.46\% & 0.0376 & 0.0741 & 0.0923 & 3.432 & 21.65 & 0.7323 & 24.26 & 2.976\\
        OmniH2O\cite{he2024omnih2o} & $15.0s$ & 94.08\% & 0.0361 & 0.0506 & 0.0914 & 3.267 & 19.86 & \textbf{0.6972} & 18.49 & 1.680\\
        Mobile-Television\cite{lu2024mobile} & $15.0s$ & 85.79\% & 0.0354 & 0.0513 & 0.0923 & 3.831 & 15.43 & 0.8775 & 11.67 & 1.677\\
        NMPC+WBC\cite{grandia2023perceptive} & $35.0s$ & 89.60\% & 0.0278 & 0.0437 & 0.0857 & 2.938 & 14.98 & - & 11.47 & -\\
        Ours (TOP) & $40.5s$ & \textbf{95.30\%} & \textbf{0.0269} & \textbf{0.0270} & \textbf{0.0827} & \textbf{2.729} & \textbf{13.45} & 0.8592 & \textbf{10.41} & \textbf{1.601}\\
        \midrule
        \rowcolor{gray!30} \multicolumn{11}{l}{\textbf{Ablation Results}}\\
        \midrule
        Ours w/o TOP $\Delta t = 0.01s$ & $\bm{15.0s}$ & 82.43\% & 0.0301 & 0.0541 & 0.0962 & 4.438 & 16.87 & 0.9329 & 12.04 & 1.738\\
        Ours w/o TOP $\Delta t = 0.03s$ & $45.0s$ & 87.27\% & 0.0303 & 0.0434 & \textbf{0.0810} & 3.797 & 16.52 & 0.9473 & 11.26 & 1.647\\
        Ours w/o TOP $\Delta t = 0.05s$ & $75.0s$ & 92.41\% & 0.0299 & 0.0362 & 0.0813 & 3.573 & 15.79 & 0.9340 & \textbf{9.79} & \textbf{1.556}\\
        Ours w/o motion prior & $34.5s$ & 89.33\% & 0.0304 & \textbf{0.0313} & 0.0824 & 3.682 & 14.76 & 0.9130 & 10.92 & 1.640\\
        Ours w/o acting chunking & $28.5s$ & \textbf{93.16\%} & \textbf{0.0288} & 0.0353 & 0.0817 & \textbf{3.275} & \textbf{14.28} & \textbf{0.8835} & 11.51 & 1.659 \\
        \bottomrule
    \end{tabular}
    }
    \caption{Comparative results and Ablation results. We execute more than 10000 motion clips from dataset $\mathcal{M}$ and unseen dataset $\mathcal{T}$, and report their time cost, success rate, mean metrics. Mean Metrics are only calculated based on the data of the robot standing in place, and the data of robot taking steps and falls will not be included in the calculation. }
    \label{tab:simulation_results}
    \vspace{-10pt}
\end{table*}

\begin{itemize}
    \item \textbf{Fixed Root (reference)}: This baseline is that we fixed the root of robot, and directly execute the upper-body motions, which means the motions can be perfectly executed with the highest control accuracy.

    \item \textbf{Exbody}\cite{cheng2024expressive}: This baseline uses whole-body RL policy with the tracking rewards to control whole-body joints. 
    
    \item \textbf{OmniH2O}\cite{he2024omnih2o}: This baseline uses whole-body RL policy with the tracking rewards to control whole-body joints, and similar to the Exbody in terms of reward settings and RL training.
    
    \item \textbf{Mobile-Television}\cite{lu2024mobile}: This baseline uses decoupling controller with motion prior, which is similar to the \textbf{Ours w/o TOP $\Delta t = 0.01s$}. The code is not open source, so we implemented it on our own robot based on the reference paper.
    
    \item \textbf{NMPC+WBC}\cite{grandia2023perceptive}: This baseline uses a non-linear Model Predictive Control (NMPC) combined with Whole-Body Control (WBC). And we slow down the target motions of NMPC for better comparison with our TOP method.
    
    \item \textbf{Ours (TOP)}: Our method of time optimization policy with balance lower-body RL policy. 
    
    \item \textbf{Ours w/o TOP $\Delta t = 0.01s/0.03s/0.05s$}: These baselines only use the balance lower-body RL policy without TOP. The timestamp between motion clips is fixed as $\Delta t$.

    \item \textbf{Ours w/o motion priors}: This baseline uses TOP method but without the motion priors. The VAE module is entirely removed and the RL and TOP policies only based on the current motion state without any latent code.

    \item \textbf{Ours w/o acting chunking}: Our method of TOP, but the time optimization policy only predicts single $\Delta t_t$ not a sequence of $\Delta t_t^{seq}$. 

\end{itemize}

The metrics are as follows:

\begin{itemize} 

\item[-] \textbf{Time Cost}: Average time cost of per 1500 motion clips (15s of origin data), which shows the time efficiency of executing the target motions.

\item[-] \textbf{Success Rate}: We define that the robot executes the upper body motion without falling, this motion clip will be labeled as successful motion. 

\item[-] \textbf{Precision}: upper joint position error $\mathbf{E^{\text{upper}}_{\text{jpe}}}$, upper end effector position error in world frame $\mathbf{E^{\text{upper}}_{\text{eepe}}}$, upper end effort orientation error in world frame $\mathbf{E^{\text{upper}}_{\text{eeoe}}}$.

\item[-] \textbf{Stability}: projected gravity $\mathbf{E_g}$, lower joint acceleration $\mathbf{E^{\text{lower}}_{\text{acc}}} $, lower action difference $\mathbf{E^{\text{lower}}_{\text{action}}}$.

\item[-] \textbf{Smoothness}: upper joint acceleration $\mathbf{E^{\text{upper}}_{\text{acc}}}$, upper action difference $\mathbf{E^{\text{upper}}_{\text{action}}}$.
\end{itemize}

\begin{figure*}[ht]
    \centering
    \setlength{\abovecaptionskip}{-0.1cm}
    \setlength{\belowcaptionskip}{-0.2cm}
	\subfigure{
		\begin{minipage}[t]{0.98\textwidth}
			\centering
			\includegraphics[width=1\textwidth]{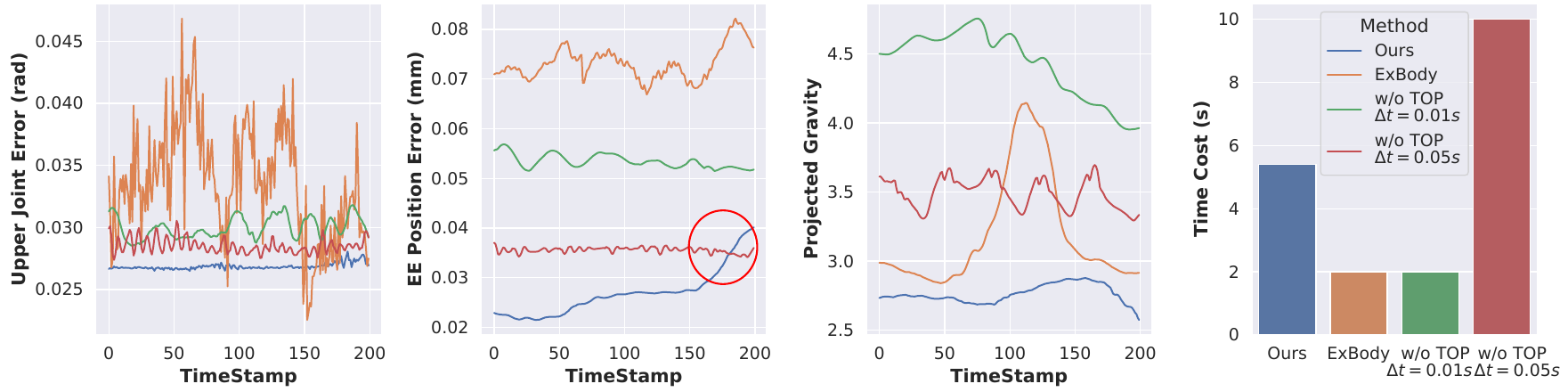}
		\end{minipage}
	}
    
    \caption{Evaluation of precision, stability, and time efficiency. We only plot results for Exbody and w/o TOP ($\Delta t = 0.01s$), since Exbody and OmniH2O show very similar behavior, while Mobile-Television is comparable to w/o TOP ($\Delta t = 0.01s$). Red circle: Because upper-body motions can occasionally have a minor impact on whole-body balance, our method prefers to execute the motion at its original speed rather than slow it down. This may slightly increase the EE Position Error but represents a dynamic balance between precision, stability, and time efficiency.}
    \label{fig::figure_evaluate}
    \vspace{-10pt}
\end{figure*}

\noindent \textbf{Analysis of TOP methods.} The results are shown in \Cref{tab:simulation_results}. It is evident that, in comparison to the Exbody or OmniH2O method, the proposed method enhances the tracking accuracy of the joint position and end effector position, but has worse time efficiency. We would like to clarify, this because our method will execute the upper-body motion directly, while the Exbody or OmniH2O method involves executing upper-body motions through an RL policy, which changes the amplitude of the robot's upper-body motions to maintain balance. That is why these method performs worse in precision. And our method will consider the impact of momentum changes that will slow down the motion, leading to more time cost. 

\begin{figure}[htbp]
      \vspace{-5pt}
      \setlength{\abovecaptionskip}{-0.1cm}
      \setlength{\belowcaptionskip}{-0.1cm}
        \centering
	\subfigure{
        \rotatebox{90}{\scriptsize{~~~~~~~TOP~~~~~~~~~~~~~~w/o TOP}}
		\begin{minipage}[t]{0.45\textwidth}
			\centering
			\includegraphics[width=1\textwidth]{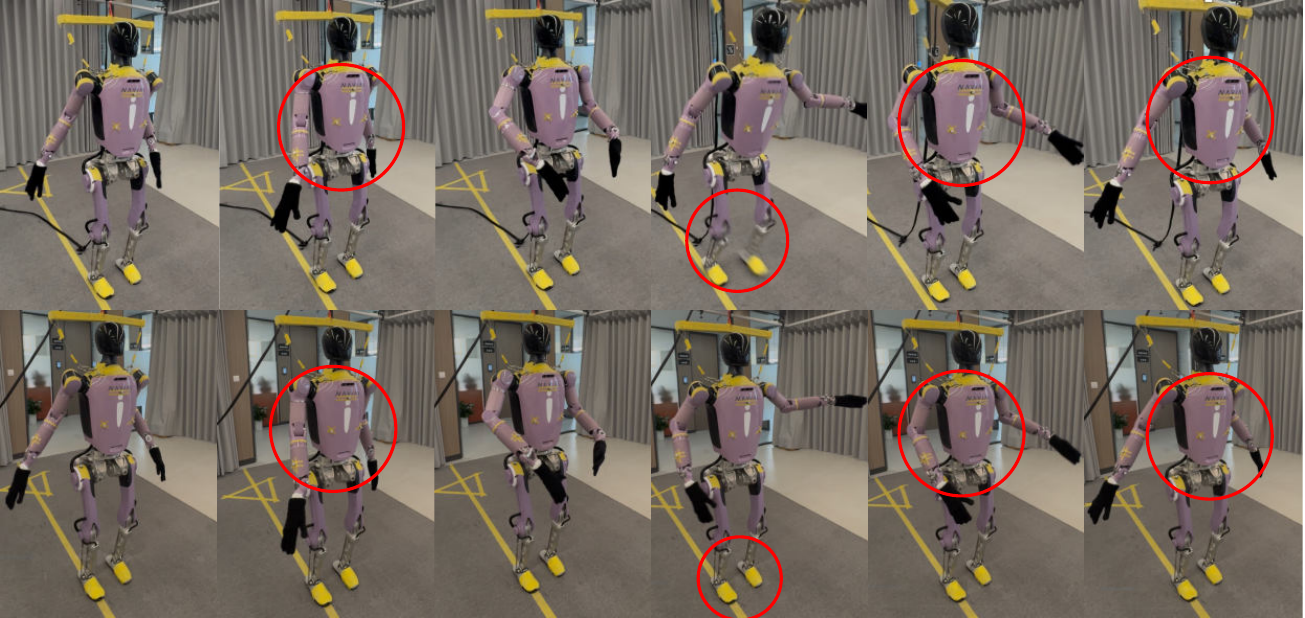}
		\end{minipage}
	}
    \caption{\textbf{TOP Performance.} The result shows that TOP can improve the stability and reduce the occurrence of robot taking a step or falling in real world.}
    \label{fig::top_evaluate}
    \vspace{-10pt}
\end{figure}

\noindent Considering the ablation results, as shown in the \Cref{fig::top_evaluate}, our method helps reduce the balance burden and unpredicted inference caused by the lower-body, which enhances both the balance stability and the precision of upper-body motions. And with the help of TOP algorithm, we have achieved a good balance between time efficiency, and accuracy, demonstrating the outperfomance of our methods. Compare to \textbf{Ours w/o TOP $\Delta t = 0.01s/0.03s/0.05s$}, our method can automatically choose the appropriate $\Delta t \in [0.01s, 0.1s]$ in different situations: when motions seriously affect accuracy and balance, our algorithm will give a larger $\Delta t$ even more than $0.05s$ to improve stability and precision; conversely, when motions are easy enough, a smaller $\Delta t$ is chosen to maximize time efficiency. Although our methods take more time to complete motions compared to \textbf{Ours w/o TOP $\Delta t = 0.01s$} and \textbf{Mobile-Television}, but we improve both precision and stability significantly. This adaptive TOP strategy results in improved overall precision and stability, while achieving higher average time efficiency. The performance curves of different methods over time are presented in Fig.~\ref{fig::figure_evaluate}, demonstrating the precision, stability and time efficiency of motion clips.

\begin{figure}[htbp]
      \vspace{-5pt}
      \setlength{\abovecaptionskip}{-0.1cm}
      \setlength{\belowcaptionskip}{-0.1cm}
        \centering
	\subfigure{
		\begin{minipage}[t]{0.48\textwidth}
			\centering
			\includegraphics[width=1\textwidth]{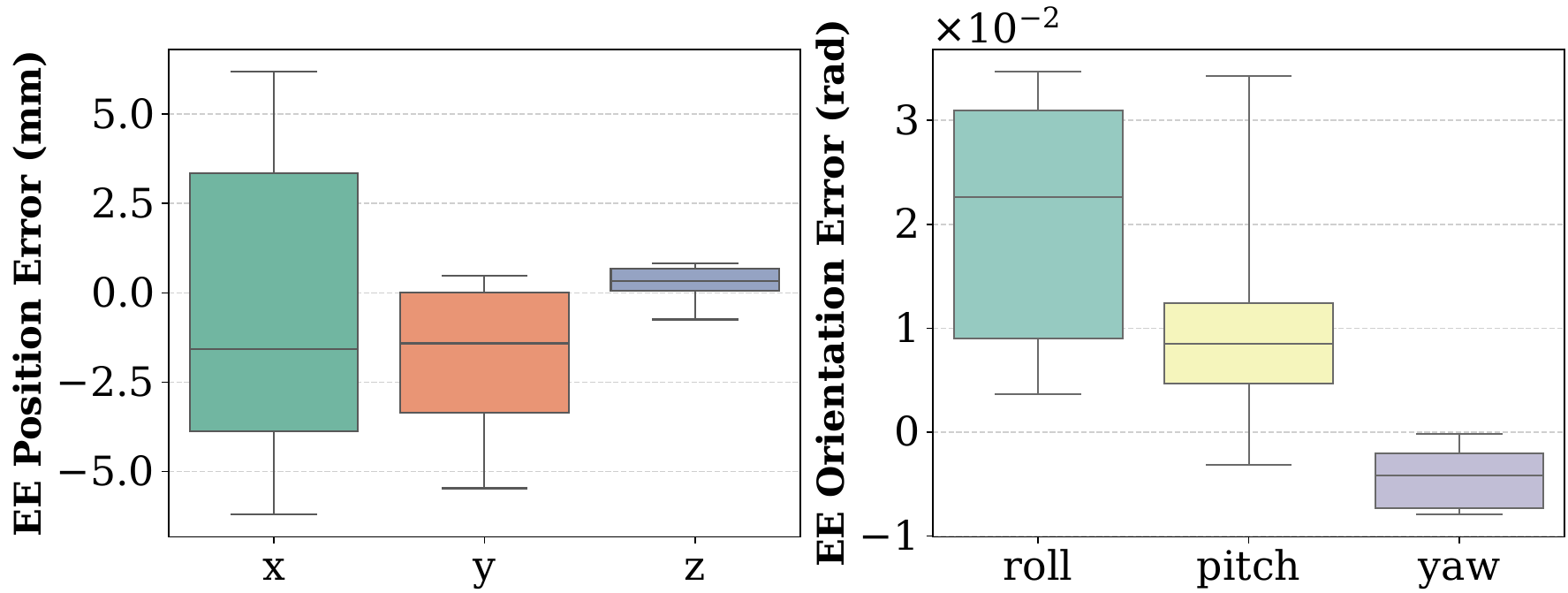}
		\end{minipage}
	}
    \caption{\textbf{Evaluate of accuracy.} EE position and orientation error of our method. The motions are from $\mathcal{M}$ which includes the motion space and speed of the vast majority of manipulation tasks, and can well measure the accuracy of our algorithm in manipulation tasks. }
    \label{fig::figure_box}
    \vspace{-10pt}
\end{figure}

\noindent We also measure the real-world base and end effector poses within accurate motion capture device, which proves that our method can still maintain balance and demonstrate high accuracy considering the time efficiency of motions. Together with the robust lower-body controller, the adaptive TOP module plays a complementary role, and both are essential to achieve the observed improvements in overall performance. The quantitative results are shown in \Cref{table::real_evaluate} and \Cref{fig::figure_box}.

\begin{table}[htbp]
\vspace{-5pt}
\centering
\caption{Real-world performance of several motions.}
\label{table::real_evaluate}
\renewcommand{\arraystretch}{1.2} 
\setlength{\tabcolsep}{6pt}       
\begin{threeparttable}
\resizebox{0.48\textwidth}{!}{
\begin{tabular}{llcccccc}
    \toprule
    \rowcolor{red!30}
    \multicolumn{2}{c}{\textbf{Part}} & \textbf{$\mathbf{x}$ (mm)} & \textbf{$\mathbf{y}$ (mm)} & \textbf{$\mathbf{z}$ (mm)} & 
    \textbf{Roll (rad)} & \textbf{Pitch (rad)} & \textbf{Yaw (rad)} \\ 
    \midrule
    \multirow{2}{*}{$\mathcal{M}$} 
        & Base Error & 1.654 & 0.461 & 0.123 & 0.0091 & 0.0114 & 0.0045 \\
    & EE Error   & 5.068 & 2.192 & 0.560 & 0.0213 & 0.0138 & 0.0048 \\
    \midrule
    \multirow{2}{*}{$\mathcal{T}$} 
        & Base Error & 29.232 & 2.485 & 1.169 & 0.0163 & 0.0119 & 0.0032 \\
    & EE Error   & 33.735 & 3.049 & 1.491 & 0.0502 & 0.0127 & 0.0103 \\
    \bottomrule
\end{tabular}
}
\begin{tablenotes}
\footnotesize
\begin{minipage}[t]{0.45\textwidth}
\item[1] The errors are averaged over twenty different motions.
\item[2] Motions from $\mathcal{M}$ have smaller amplitude and slower speed of the motions than motions from $\mathcal{T}$.
\item[3] It is evident that the error of the base accounts for the majority of the  end-effector pose error.
\end{minipage}
\end{tablenotes}
\end{threeparttable}
\vspace{-10pt}
\end{table}


\subsection{Manipulation Experiments} 
\begin{figure}[ht]
      \vspace{0pt}
      \setlength{\abovecaptionskip}{-0.1cm}
      \setlength{\belowcaptionskip}{-0.1cm}
      \subfigure{
		\begin{minipage}[t]{0.47\textwidth}
			\centering
			\includegraphics[width=1\textwidth]{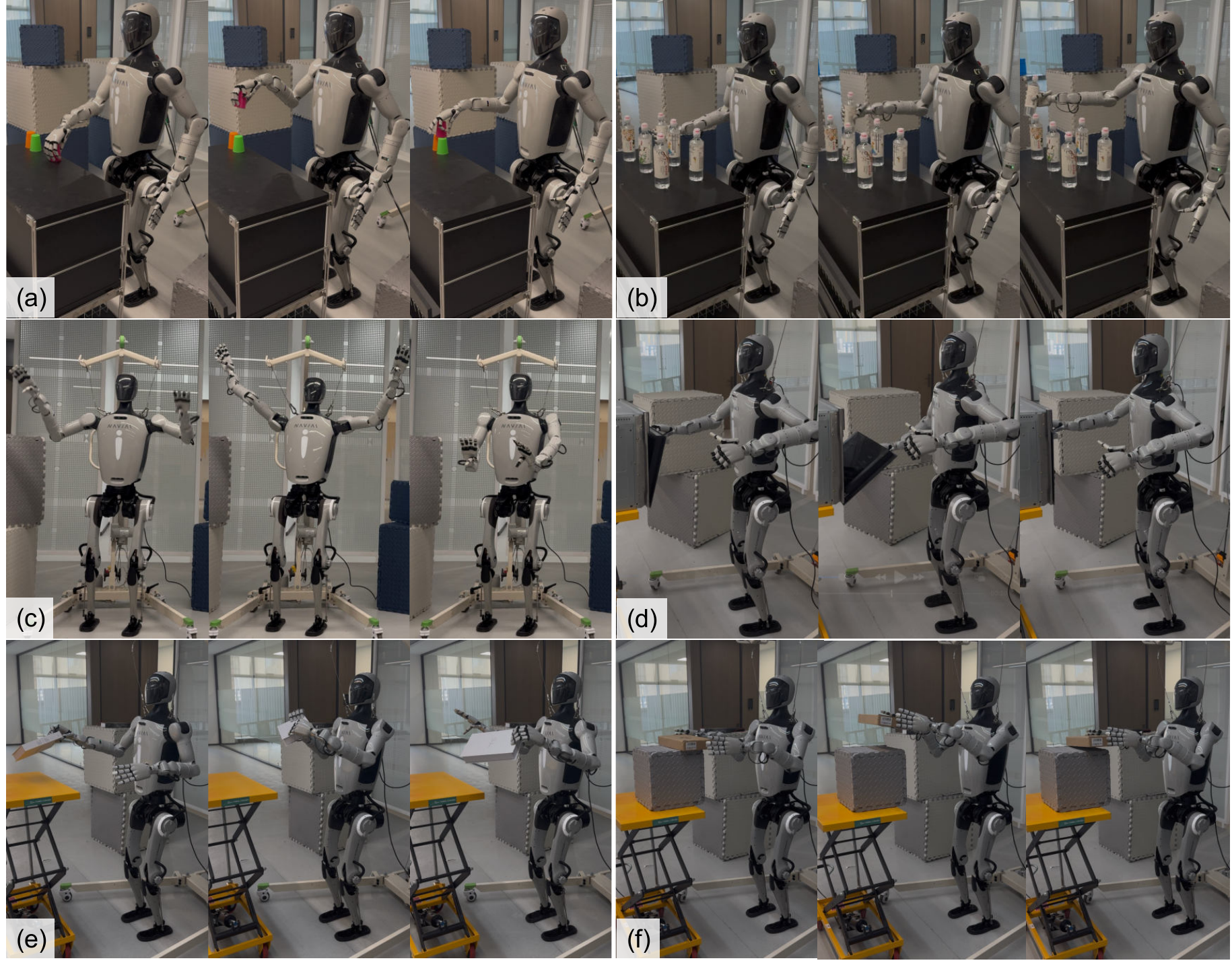}
		\end{minipage}
	}
    \caption{\textbf{Manipulation tasks.} (a) grab the cup and put it onto cups, (b) take and deliver a bottle of water, (c) dance with arms but slow down the original motions while the lower body maintains balance, (d) open the oven, (e) Take the book and pass it from left hand to right hand, (f) carry a box.}
    \label{fig::teleop_evaluate}
    \vspace{-10pt}
\end{figure}

We evaluate our approach on manipulation tasks that require precision and robustness simultaneously, and we also use teleoperation combined with TOP for real-time motion slowing. In simple manipulation tasks, like grab a cup, policy tends to maintain the original speed to get higher time efficiency. When the amplitude and speed of the robot's motion affect balance, like dance with arms, it will actively slow down the motion, which can reduce the impact of momentum on the robot to control the robot stably and accurately. The results are shown in \Cref{fig::teleop_evaluate}.

\section{CONCLUSIONS}
In this paper, we propose a novel framework with the TOP method, which improves lower-body stability and upper-body precision by optimizing the timestamps of motion clips. By integrating a robust lower-body policy with precise upper-body tracking, and leveraging a VAE-based motion prior to capture upper-body features, our method achieves 95.30\% success rates in experiments, as reported in \cref{tab:simulation_results}, demonstrating stable and accurate standing manipulation performance. However, current motions remain somewhat rigid, for instance, the robot tends to tilt backward rather than twisting its hips. In future work, we plan to incorporate motion generation modules to enable more natural standing adjustments and further enhance balance control. Moreover, we aim to explore adaptive task-oriented time optimization policy to extend our application in real-world tasks.

\addtolength{\textheight}{0.2cm}   








\vspace{-5pt}
\printbibliography


\enlargethispage{\baselineskip}


\section{APPENDIX}

\subsection{Motion Alignment And Coupling Effects}
In principle, upper- and lower-body motions are inherently coupled through momentum transfer and base motion feedback. Fully ignoring this coupling would risk infeasible or unstable motions. In our design, however, the decoupling is a practical modular abstraction rather than a rigid separation. Upper-body motion is compressed by a VAE-based motion prior and tracked by a PD controller that outputs desired joint torques to follow the reference trajectories with high precision, while lower-body RL policy receives real-time observations of the robot state, and the desired upper-body motion embedding ($\bm{z_t}$) and motion clip ($\bm{m_t}$), compensating for the resulting momentum changes by dynamically regulating base drift and joint torques.

Therefore, while the modules are trained in a decoupled manner, their execution forms a closed-loop whole-body control system. The RL policy implicitly compensates for momentum transfer from the upper-body by dynamically adjusting lower-body joint torques and contact forces to maintain balance. This design principle is consistent with established whole-body control frameworks\cite{lu2024mobile} where manipulation and locomotion are hierarchically coordinated. Our reward structure explicitly penalizes excessive base drift and velocity deviations, enforcing feasible whole-body dynamics, as shown in the \cref{table_param}.

As shown in \cref{tab:simulation_results}, compared to baseline methods (Exbody\cite{cheng2024expressive}, OmniH2O\cite{he2024omnih2o}) that rely on end-to-end whole-body RL tracking, our hybrid modular strategy achieves higher tracking precision, base stability, and overall motion success rate. One detail worth clarifying is the slightly higher upper-body acceleration error observed for our method (TOP) relative to the baseline (\textbf{Ours w/o TOP $\Delta t = 0.05s$}). This effect stems from our time optimization policy, which adaptively adjusts the execution speed of the upper-body trajectory in response to the robot’s real-time dynamic state. Such non-uniform time scaling naturally introduces local variations in velocity and acceleration, resulting in occasional mild phase lags or higher acceleration spikes when the speed changes rapidly. In contrast, the baseline uses a fixed time step for the entire trajectory, which reduces local acceleration fluctuations but sacrifices dynamic adaptability and time efficiency.

In summary, although the TOP policy dynamically adjusts upper-body timestamps, the lower-body actions are not static but adapt in real-time to ensure whole-body consistency. The empirical results and reward design confirm that our approach avoids dynamic misalignment and preserves stable and feasible behavior.

\subsection{Quantitative Regression Analysis}

\cref{fig::figure_evaluate} qualitatively illustrated the great balance between precision, stability, and time efficiency. To provide a more rigorous quantitative characterization of these interdependencies, we additionally conducted a regression and Pareto analysis. 

Specifically:
\begin{itemize}
\item Multiple linear and polynomial regressions were applied to model the relationship between time efficiency ($\Delta t$), base stability (projected gravity), and precision (end-effector position error).
\item An empirical Pareto front\cite{aydin2020computational} was computed to reveal the boundary of non-dominated solutions, illustrating the inherent trade-offs.
\end{itemize}

\begin{figure*}[htbp]
    \centering
    \setlength{\abovecaptionskip}{-0.1cm}
    \setlength{\belowcaptionskip}{-0.2cm}
    \includegraphics[width=0.9\linewidth]{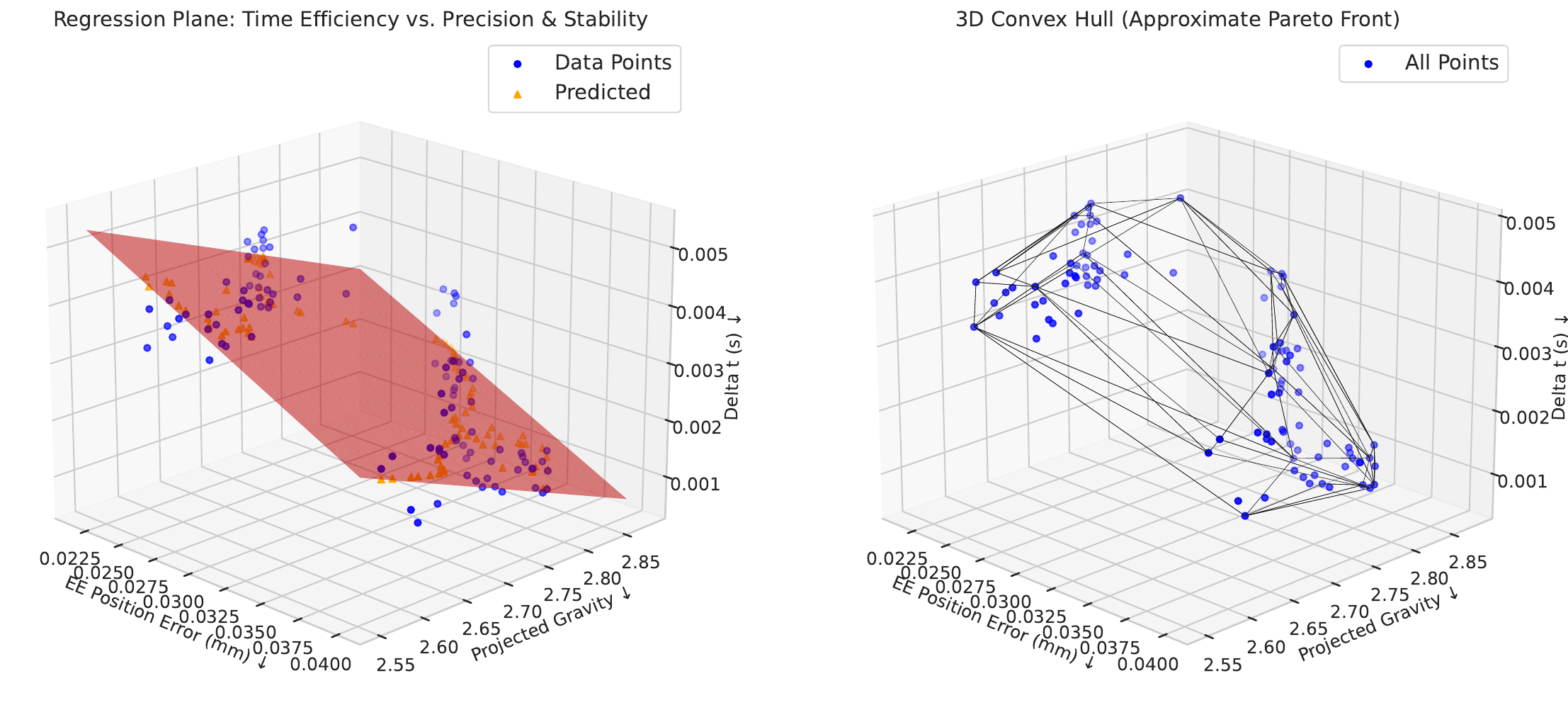}
    \caption{\textbf{Regression and Pareto front analysis.} The figure shows the multi-objective trade-off among time, stability, and precision, with the fitted regression surface and the extracted Pareto optimal front (right).}
    \label{fig:regression_pareto}
    \vspace{-10pt}
\end{figure*}

\cref{fig:regression_pareto} shows the updated results. Each scatter point represents an individual upper-body motion trial with its measured $\Delta t$, projected gravity, and pose error. The fitted regression surface captures the trend that smaller $\Delta t$ generally leads to slightly higher projected gravity and pose error due to increased momentum transfer from rapid upper-body motions. The highlighted Pareto front marks the boundary where improvement in one metric cannot be achieved without compromising at least one of the others.

This analysis confirms that higher end-effector accuracy is not independent but is coupled to base stability and $\Delta t$. Larger $\Delta t$ implies slower motions, allowing the lower-body controller more time to absorb upper-body momentum, resulting in lower projected gravity and reduced pose error. Conversely, executing motions faster increases local accelerations, slightly degrading base stability and precision.

\subsection{Robustness Evaluation}

\begin{table}[h]
\centering
\vspace{-5pt}
\caption{Domain Randomization Settings}
\label{tab:domain_rand}
\resizebox{0.47\textwidth}{!}{
\begin{tabular}{ll}
\toprule
\textbf{Term} & \textbf{Value} \\
\midrule
\multicolumn{2}{l}{\textbf{Observation}} \\
Joint position noise & $\mathcal{U}(-0.1, 0.1)$ rad \\
Joint velocity noise & $\mathcal{U}(-1.0, 1.0)$ rad/s \\
Angular velocity noise & $\mathcal{U}(-0.1, 0.1)$ rad/s \\
Projected gravity noise & $\mathcal{U}(-0.05, 0.05)$ \\
\midrule
\multicolumn{2}{l}{\textbf{Dynamics}} \\
Base COM displacement & $\mathcal{U}(-0.05, 0.05)$ m \\
Friction factor & $\mathcal{U}(0.35, 1.25)$ \\
Link Inertia & $\mathcal{U}(0.8, 1.2) \times $ default kg $\cdot$ m$^2$ \\
Added base mass & $\mathcal{U}(-5.0, 5.0)$ + default kg \\
Added link mass & $\mathcal{U}(0.8, 1.2) \times $ default kg \\
Added hand mass & $\mathcal{U}(-0.5, 1.5)$ + default kg \\
Payload mass & $\mathcal{U}(-1.0, 3.5)$ + default kg \\
Motor P gain & $\mathcal{U}(0.8, 1.2) \times $ default \\
Motor D gain & $\mathcal{U}(0.8, 1.2) \times $ default \\
Motor strength & $\mathcal{U}(0.9, 1.1) \times $ default N $\cdot$ m\\
Motor Damping & $\mathcal{U}(0.3, 4.0) $ N $\cdot$ s \\
Motor Delay & $\mathcal{U}(0, 10) $ ms \\
\midrule
\multicolumn{2}{l}{\textbf{Episode}} \\
Push robots & interval = 5s, $v_{xy} =$ 1.0m/s, $\omega$ = 0.4rad/s \\
Disturbance robots & interval = 8s, $\mathcal{U}(-30.0, 30.0)$ N $\cdot$ m \\
Initial robot base position xy & $\mathcal{U}(-1.0, 1.0)$ m \\
Initial robot base velocity & $\mathcal{U}(-0.1, 0.1)$ m/s \\
\bottomrule
\end{tabular}
}
\vspace{-10pt}
\end{table}

Although our VAE focuses on learning compact latent representations of human-like motion purely from joint position and velocity data, without directly incorporating dynamic constraints (e.g., torque limits, inertia matrices). However, the final execution relies on a downstream reinforcement learning (RL) policy, and we incorporate extensive domain randomization during RL training. As summarized in \cref{tab:domain_rand}, this includes random variations in payload mass, link inertia, base mass, motor properties, friction, and observation noise. In addition, randomized external pushes and initial state perturbations are applied to expose the policy to diverse disturbance scenarios.

To empirically verify this design, we conducted an additional load-adaptability test, in which the robot performed the same manipulation task while grasping objects of increasing mass. \cref{tab::load_adaptability} reports the results. The task success rate and end-effector pose error remain stable across different payloads, while only a slight increase in the projected gravity term is observed. This demonstrates that our RL controller compensates for the dynamic variations introduced by heavier payloads, ensuring that the generated actions remain physically feasible and stable even though the upstream VAE module is purely kinematic.

\begin{table}[htbp]
\centering
\vspace{-5pt}
\caption{Payloads Adaptability Test}
\label{tab::load_adaptability}
\begin{tabular}{c|c|c|c}
\toprule
\textbf{Payload} & \textbf{Success Rate} $\uparrow$ & \textbf{EE Pose Error} $\downarrow$ & \textbf{Projected Gravity} $\downarrow$ \\
\midrule
0.5 kg & 95.2\% & 28.5 mm & 2.731 \\
1.0 kg & 94.4\% & 30.1 mm & 2.857 \\
3.0 kg & 92.7\% & 31.8 mm & 3.142 \\
\bottomrule
\end{tabular}
\vspace{-10pt}
\end{table}

\begin{figure}[htbp]
    \centering
    \vspace{-5pt}
    \includegraphics[width=1.0\linewidth]{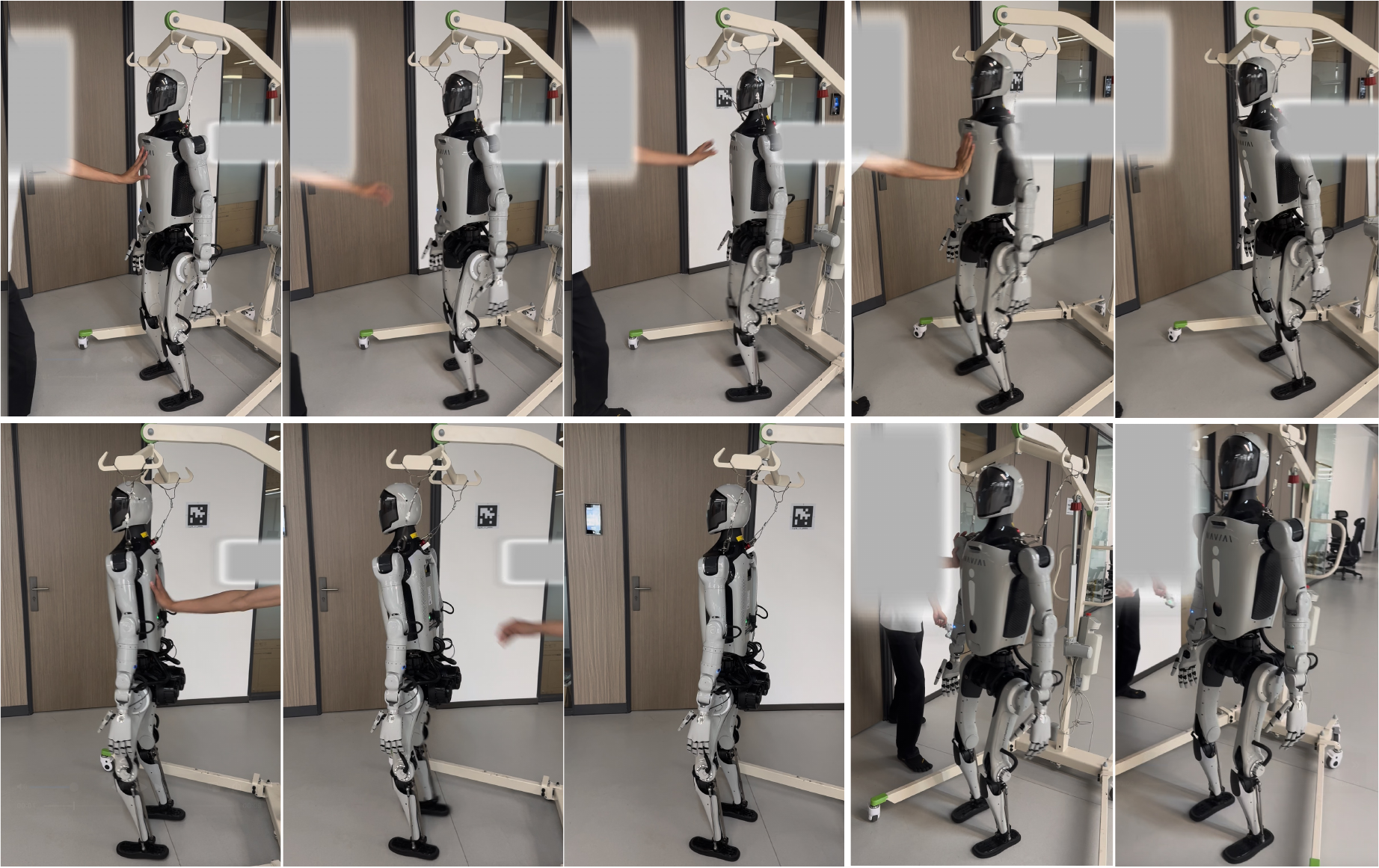}
    \caption{\textbf{External push disturbance tests.} The figure illustrates the robot can recover to balance while disturbed by external pushes from different directions.}
    \label{fig::push_robot}
    \vspace{-10pt}
\end{figure}

Also, we add more experiments about push recovery of our method in real-world robot. \cref{fig::push_robot} shows external disturbance tests of our policy. We designed scenarios where the robot is intentionally disturbed by external pushes from different directions. 

\begin{figure}[h]
    \centering
    \vspace{-0pt}
    \includegraphics[width=1.0\linewidth]{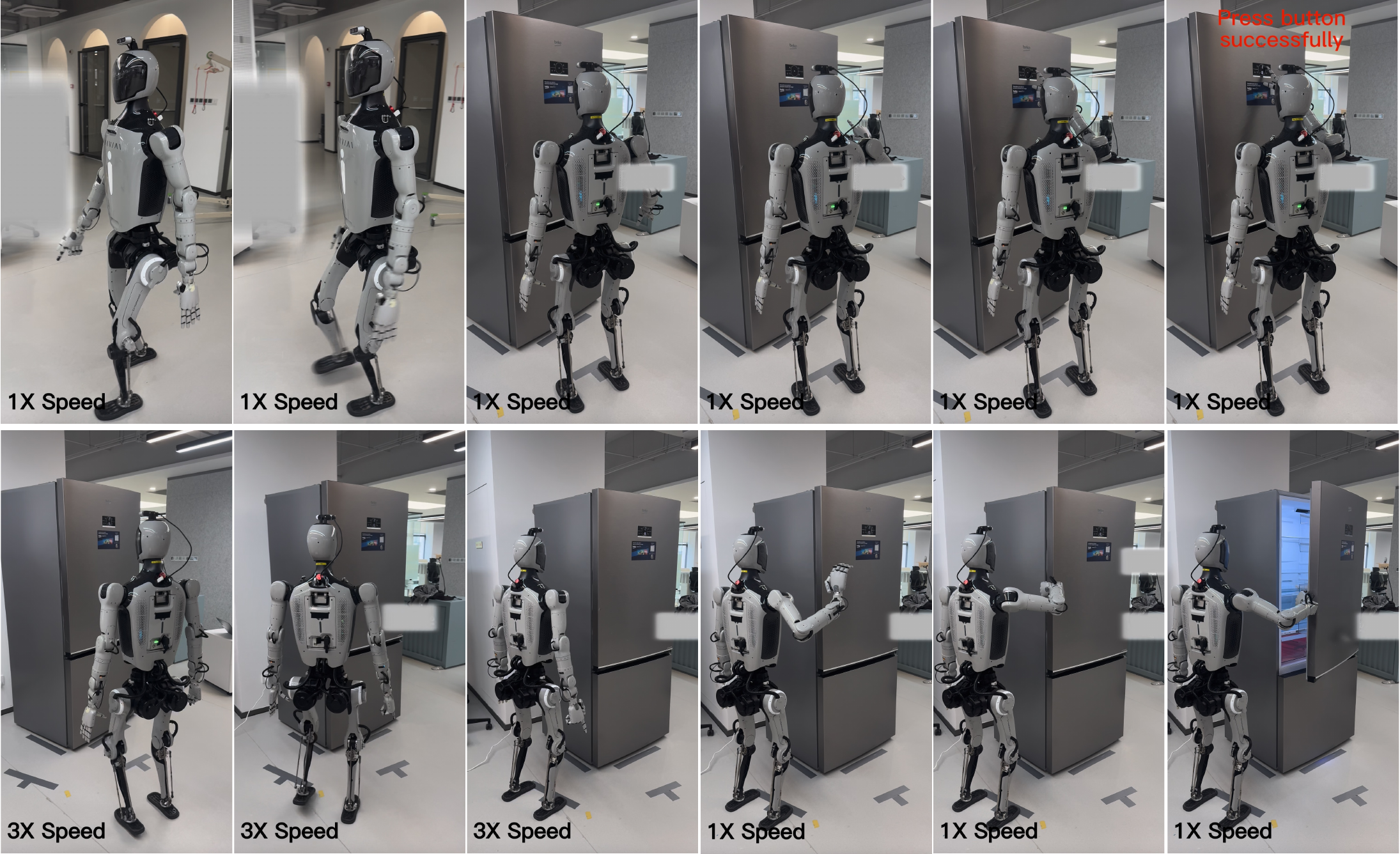}
    \caption{\textbf{Loco-manipulation scenarios.} The top figures show the robot executing a loco-button-pressing task. The bottom figures shows the robot executing a loco-open-fridge task.}
    \label{fig::loco_mani}
    \vspace{-10pt}
\end{figure}

Moreover, we have conducted additional experiments that extend beyond static standing manipulation. We now demonstrate the effectiveness and robustness of our Time Optimization Policy (TOP) when the robot performs performs loco-manipulation tasks. \cref{fig::loco_mani} shows the mobile operation scenarios of our TOP policy. In these new test cases, the humanoid robot executes upper-body tasks and omnidirectional walking. This demonstrates that the proposed decoupled yet closed-loop framework generalizes well to locomotion plus manipulation tasks.  

To make our results fully transparent, we have released all the demonstration videos on our project page \href{https://anonymous.4open.science/w/top-258F/}{\textcolor{mypink}{https://anonymous.4open.science/w/top-258F/}}.

\subsection{Measurement of Manipulation Tasks}

\begin{figure}[h]
\centering
\vspace{-5pt}
\includegraphics[width=0.95\linewidth]{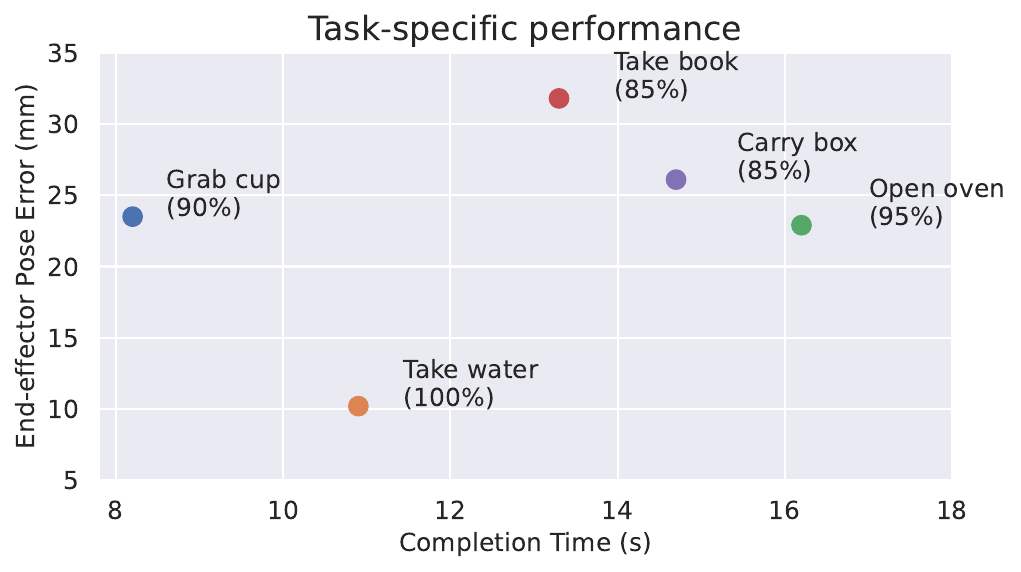}
\caption{\textbf{Comparison between different manipulation tasks.} Where each point represents a single task annotated by its success rate. The closer the data points are to the lower left corner, the lower the end-effector pose error and the shorter the task completion time, indicating better time efficiency and precision. This trend demonstrates how our method maintains high success rates while balancing precision and time efficiency across tasks of varying complexity. }
\label{fig::manipulation_tasks}
\vspace{-10pt}
\end{figure}

To complement the qualitative demonstrations in \cref{fig::teleop_evaluate}, we report detailed quantitative performance metrics for each manipulation scenario. 
\cref{fig::manipulation_tasks} presents the \textbf{success rate}, \textbf{task completion time}, and \textbf{end-effector pose error} (EE pose error) for five representative tasks: grabbing a cup, taking water, opening an oven, taking a book, and carrying a box. These metrics were obtained by repeating the real-world experiments multiple times under the same conditions.

The results demonstrate that our framework achieves high success rates and acceptable completion times while maintaining reasonable EE pose errors. For more challenging tasks such as \textit{take book} and \textit{carry box}, slightly higher pose errors are observed due to the increased complexity of manual operation and shifting loads, which introduce additional vibrations and tracking challenges on the real robot. Despite these challenges, the success rates remain competitive (around 85\%), showcasing the robustness of our method in realistic settings.

To provide more fine-grained, task-specific precision metrics as requested by the reviewer, we additionally measured detailed indicators such as handle rotation angle error and final placement offsets using a millimeter ruler, due to the difficulty of motion capture device handling the reflective objects. These results are summarized in \cref{tab::task_specific_metrics}. For example, the oven handle rotation error is kept below 5°, and object placement offsets remain within a few centimeters, verifying that the system achieves both reliable general pose tracking and practical task-level accuracy.

\begin{table}[htbp]
\centering
\caption{Detailed Task-Specific Metrics}
\label{tab::task_specific_metrics}
\begin{tabular}{lcc}
\toprule
\textbf{Task} & \textbf{Key Precision Metric} & \textbf{Value} \\
\midrule
Grab cup  & Cup final placement offset (mm) & 22.5 mm \\
Open oven & Handle rotation angle error (deg) & 4.8° \\
Take book & Book alignment deviation (mm) & 28.7 mm \\
Carry box & Box final placement offset (mm) & 25.3 mm \\
\bottomrule
\end{tabular}
\end{table}

Together, these results provide solid evidence that our method achieves reliable and efficient performance across diverse manipulation tasks, validating its practicality under real-world uncertainties.

\subsection{Inference Performance and Hardware Details}
To quantify the computational overhead of our system, we measured the inference performance of each core module, including the VAE encoder, the RL policy, and the TOP policy. \cref{tab:inference_performance} summarizes the average latency, device specifications, and resource utilization for both high-end GPU (RTX 4090) and edge computing (PICO-TGU4) platforms.

\begin{table}[htbp]
\centering
\vspace{-5pt}
\caption{Inference Performance in Difference Devices.}
\label{tab:inference_performance}
\resizebox{0.5\textwidth}{!}{
\begin{tabular}{lccc}
\toprule
\textbf{Module} & \textbf{Latency (ms)} & \textbf{Device} & \textbf{Resource Utilization} \\
\midrule
VAE Encoder     & 1.0121   & RTX 4090 GPU    & GPU Utilization: 4\%         \\
RL policy      & 1.7414    & RTX 4090 GPU    & GPU Utilization: 5\%         \\
TOP Policy      & 1.8623   & RTX 4090 GPU    & GPU Utilization: 8\%          \\
Total Inference & 4.6158   & RTX 4090 GPU    & Memory Usage: 4.410 GB          \\
\midrule
VAE Encoder     & 1.4621     & PICO-TGU4    & CPU Utilization: 6\%         \\
RL policy       & 1.6925      & PICO-TGU4    & CPU Utilization: 7\%         \\
TOP Policy      & 1.7267     & PICO-TGU4     & CPU Utilization: 7\%          \\
Total Inference & 4.8813     & PICO-TGU4    & Memory Usage: 5.4339 GB          \\
\bottomrule
\end{tabular}
}
\vspace{-10pt}
\end{table}

In actual deployment, all trained \texttt{.pt} models are exported to the \texttt{ONNX} format for optimized runtime inference. Since the network architectures are lightweight and do not contain complex structures such as Transformers, the inference is highly efficient and stable across platforms.

Overall, the combined inference time is approximately \textbf{5 ms}, enabling a control frequency of 100~Hz, which demonstrates that the proposed system is suitable for real-time applications.

\begin{figure}[htbp]
\centering
\vspace{-0pt}
\includegraphics[width=0.8\linewidth]{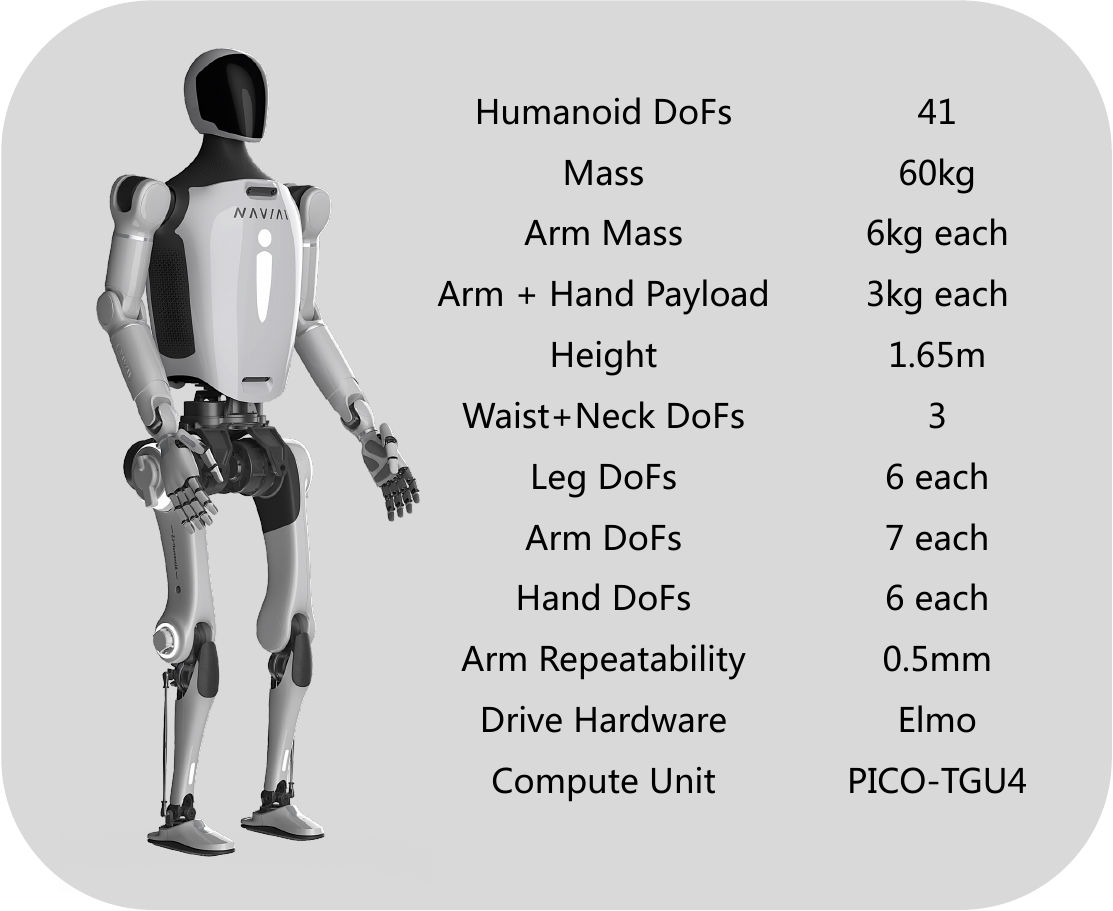}
\caption{\textbf{Hardware Details.} Our robot stands 1.65m tall, weighs 60kg, and has 41 degrees of freedom, including two 6kg arms with 7-DoF each and a payload capacity of 3kg per arm.}
\label{fig::robot_hardware}
\vspace{-10pt}
\end{figure}

In addition, to provide a clearer understanding of the experimental setup of our system, we report the detailed hardware configuration below. The humanoid robot used in our simulations and experiments stands 1.65 meters tall, weighs 60 kg, and has 41 degrees of freedom, including two 7-DoF arms (6 kg each) with a payload capacity of 3 kg per arm. \cref{fig::robot_hardware} summarizes the robot’s drive hardware and onboard computing unit configuration.

These details ensure transparency and demonstrate that our proposed system is feasible for real-time deployment on practical hardware.

\subsection{Sensitivity Analysis of the impact of k}
To improve the $\Delta t$ smoothness of the predicted timestamps and avoid jerky discrete switching, our method uses a decay coefficient~$k$ in the action chunking\cite{zhao2023learning} mechanism. The decay coefficient~$k$ defines a weight vector for aggregating recent actions when computing the final timestamp adjustment~$\Delta t$. Specifically, for a buffer of $N$ past frames, the weight for frame~$i$ is given by:
\begin{equation}
w_i = \frac{\exp(-k \cdot i)}{\sum_{j=0}^{N} \exp(-k \cdot j)}, \quad i = 0, 1, \dots, N.
\end{equation}
\noindent This assigns higher weights to recent actions and exponentially down-weights older ones. A smaller $k$ yields a flatter weight distribution (more uniform smoothing), while a larger $k$ makes the system more reactive to the latest changes. The chunking mechanism thus balances temporal consistency and reactivity by tuning~$k$.

\begin{figure}[htb]
\centering
\vspace{-5pt}
\includegraphics[width=1.0\linewidth]{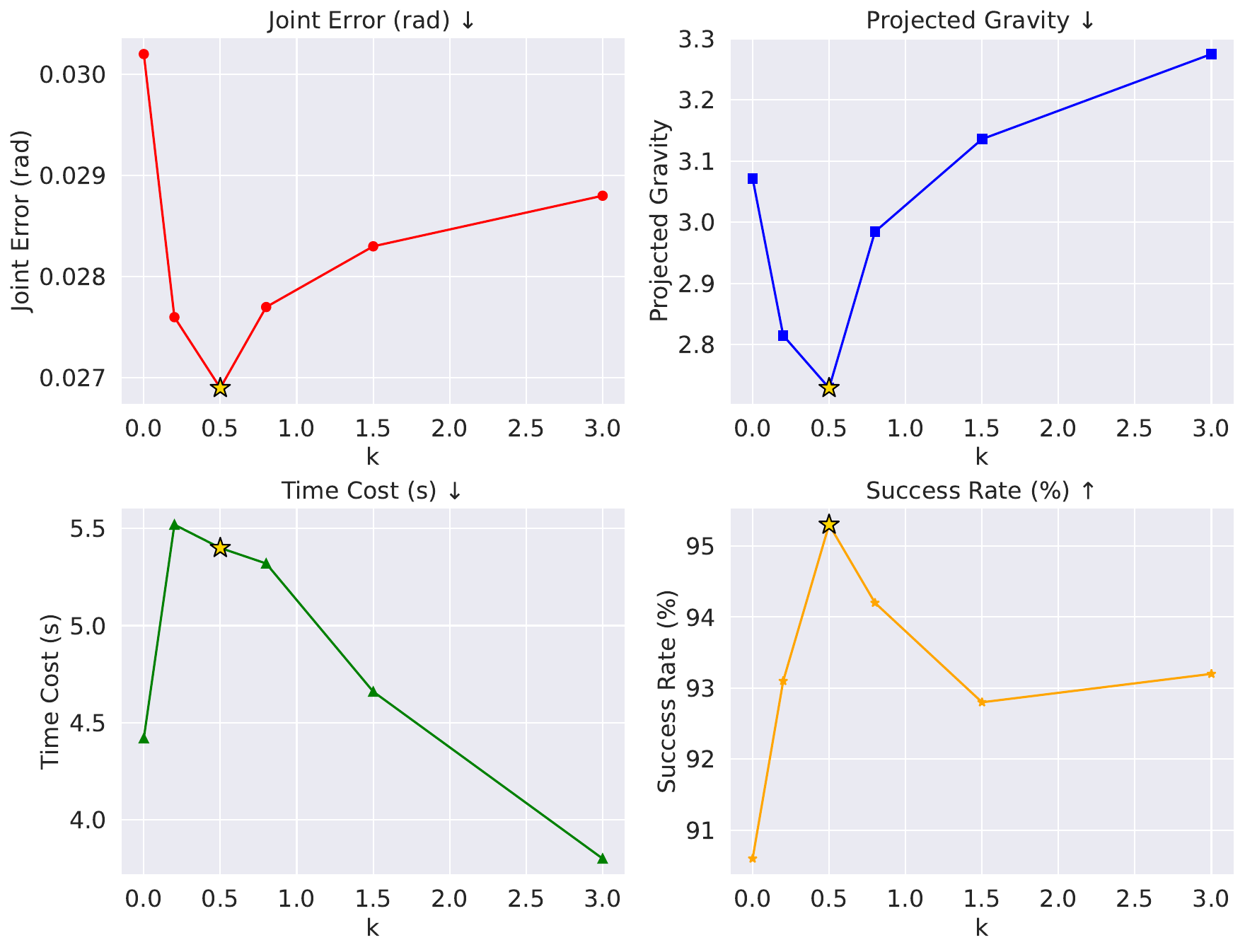}
\caption{\textbf{Sensitivity Analysis of Decay Coefficient $k$.} The plots show how $k$ affects (a) Joint Error, (b) Projected Gravity, (c) Time Cost, and (d) Success Rate. The golden point marks $k=0.5$, which achieves the best overall trade-off.}
\label{fig::k_ablation}
\vspace{-10pt}
\end{figure}

We conducted a sensitivity study by varying $k$ in ${0.0, 0.2, 0.5, 0.8, 1.5, 3.0}$ and measuring its impact on joint tracking error, base stability (projected gravity), execution time cost, and success rate. \cref{fig::k_ablation} summarize the results.

When $k$ is very small ($k=0.0, 0.2$), the weights are nearly uniform, which overly smooths the timestamps and reduces responsiveness to rapid state changes. When $k$ is too large ($k=1.5, 3.0$), the chunking emphasizes only the most recent action, losing the benefit of temporal consistency and slightly increasing error and base drift. A moderate value ($k=0.5$) achieves the best balance: it yields the lowest joint error (0.0269~rad) and the lowest projected gravity (2.7290), with a reasonable time cost and the highest task success rate (95.3\%). This empirical result supports the choice of $k=0.5$ as the default setting for our time optimization policy.

\subsection{Sensitivity Analysis of VAE parameters}

The latent code dimension and window size are critical hyperparameters for ensuring that the VAE can effectively extract meaningful motion priors. Therefore, we conducted an ablation study to evaluate how these settings affect the reconstruction quality, KL divergence loss, and overall generalization.

\cref{tab:vae_ablation} summarizes the results. A moderate window size of 30 frames achieves the lowest reconstruction error and stable KL loss on both training and test sets. Shorter window size (e.g. 15) provide insufficient temporal context for modeling complex motion patterns, which leads to poorer test performance despite similar training loss. Conversely, overly long windows size (e.g. 60) increase the temporal span to be jointly encoded, introducing noise, higher variance, and optimization difficulty, which degrades test quality due to overfitting.

\begin{table}[htbp]
\centering
\vspace{-5pt}
\caption{Sensitivity to Latent Dimension and Window Size}
\label{tab:vae_ablation}
\begin{tabular}{cc|cc|cc|c}
\toprule
\textbf{Window} & \textbf{Latent} & \multicolumn{2}{c|}{\textbf{Recon. Loss}} & \multicolumn{2}{c|}{\textbf{KLD Loss}} & \textbf{Time} \\
\textbf{Size} & \textbf{Dim} & \textbf{Train} & \textbf{Test} & \textbf{Train} & \textbf{Test} & \textbf{(hrs)} \\
\midrule
15  & 64  & 0.0261 & 0.3751 & 0.0148 & 0.0207 & 3.6 \\
30  & 64  & \textbf{0.0250} & \textbf{0.2495} & 0.0126 & 0.0143 & 4.2 \\
60  & 64  & 0.0714 & 0.6992 & \textbf{0.0081} & \textbf{0.0096} & 4.7 \\
\midrule
30  & 32  & 0.1139 & 2.7729 & \textbf{0.0086} & 7.8787 & 4.0 \\
30  & 64  & 0.0250 & \textbf{0.2495} & 0.0126 & \textbf{0.0143} & 4.2 \\
30  & 128 & \textbf{0.0238} & 0.2807 & 0.0118 & 0.0161 & 5.3 \\
\bottomrule
\end{tabular}
\vspace{-10pt}
\end{table}

Regarding latent dimension, 64 achieves the best trade-off between representation capacity and regularization. A smaller code (32) lacks sufficient capacity, resulting in high test reconstruction and KL losses and unstable latent encoding. A larger dimension (128) slightly improves training performance but increases test error and computational cost due to overfitting and redundant capacity.

\subsection{Analysis of Stability}
To provide theoretical evidence of whole-body stability, we complement our empirical results with an explicit analysis based on the Zero Moment Point (ZMP) criterion. The ZMP is defined as the point on the ground plane where the horizontal component of the moment generated by the ground reaction force becomes zero. A necessary condition for dynamic equilibrium is that the ZMP must remain inside the support polygon formed by all contact points. To generalize this idea to non-planar contact scenarios, we adopt the concept of the Zero Moment Line (ZML)~\cite{brecelj2022zero}, which extends ZMP to varying contact heights.

The ZMP is calculated by solving:
\begin{align}
    &\bm{\tau} = \bm{p}_{zmp} \times \bm{f} + \bm{\tau}_{p} \\
    &\mathcal{\bm{\dot{P}}} = M\bm{g} + \bm{f} \\
    &\mathcal{\bm{\dot{L}}} = \bm{p}_{com} \times M\bm{g} + \bm{\tau}
\end{align}

\noindent where $p_{zmp}$ is the position of ZMP, $\tau_p$ is the moment about the ZMP point, and the $f$ represents the ground reaction force. $P$, $L$, $M$, $p_{com}$ represents the linear momentum, the angular momentum, the total mass, the center of mass (CoM) position of the system, respectively, and $g = [0, 0, -g]^T $. By the definition of ZMP, we have:
\begin{equation}
    \bm{\tau}_{p, x} = \bm{\tau}_{p, y} = 0
\end{equation}

\begin{figure}[htbp]
\centering
\vspace{-5pt}
\includegraphics[width=0.3\linewidth]{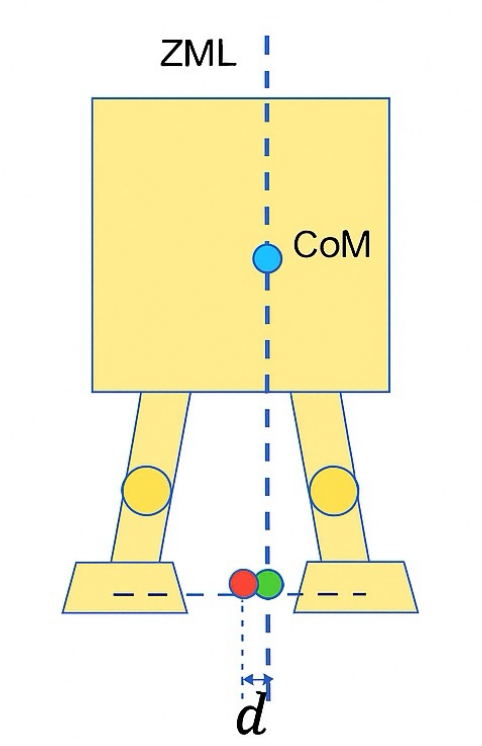}
\caption{\textbf{ZML cross the support polygon.} The red dot shows the approximate center of the support region; the green dot shows the projected ZML point. The distance $d$ quantifies stability: smaller $d$ implies better dynamic balance.}
\label{fig::ZML}
\vspace{-10pt}
\end{figure}

\noindent Solving the above equations with respect to $\bm{p}_{zmp}$ , we have
\begin{align}
    \label{zmp_equation}
    &p_{zmp, x} = \frac{Mgp_{com,x} + p_{com, z} \mathcal{\dot{P}}_x -\mathcal{\dot{L}}_y}{Mg + \mathcal{\dot{P}}_z} \\
    &p_{zmp, y} = \frac{Mgp_{com,y} + p_{com, z} \mathcal{\dot{P}}_y +\mathcal{\dot{L}}_x}{Mg + \mathcal{\dot{P}}_z}
\end{align}
\noindent By varying the height $p{com,z}$, these equations generate the ZML, whose projection must intersect the support polygon to ensure balance, as illustrated in \cref{fig::ZML}.

\begin{figure}[htbp]
\centering
\vspace{-5pt}
\includegraphics[width=1.0\linewidth]{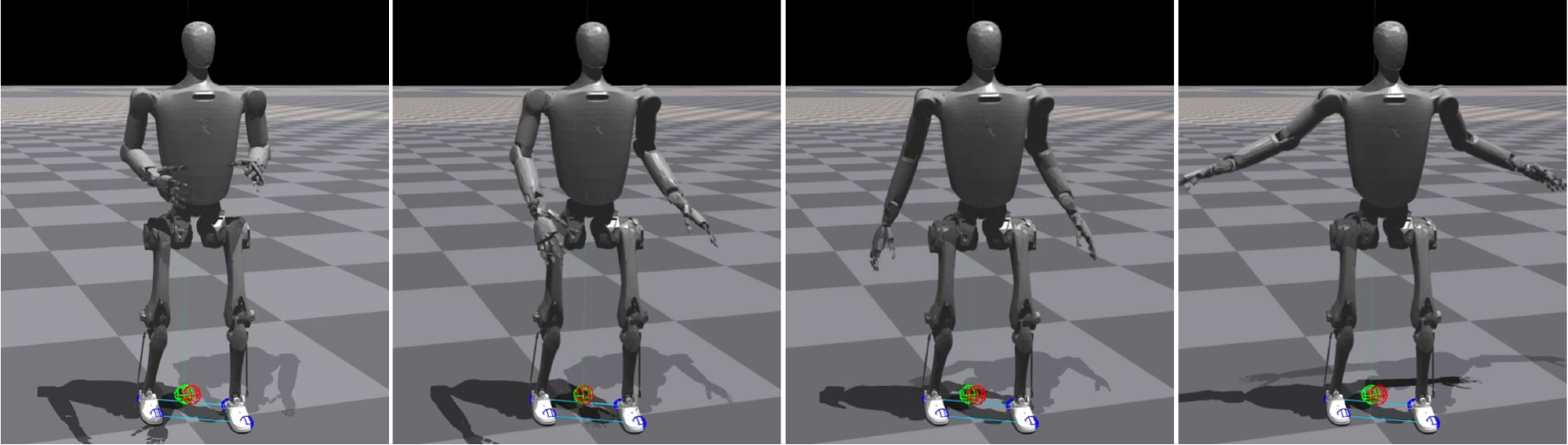}
\caption{\textbf{ZML and Support Polygon in Isaacgym.} The red point represents the center of the support polygon, and the green dot on the ZML is the projected point at the same horizontal height as the red dot. The blue points and the light blue lines are the support polygon.}
\label{fig::ZML_isaac}
\vspace{-10pt}
\end{figure}

In practice, the support polygon is computed as the convex hull of all current contact points. The ZML is computed in simulation for all time steps (\cref{fig::ZML_isaac}). We define the stability metric $d$ as:

\begin{equation}
    d = \min_{p_{csp} \in \mathcal{S}} \| p_{csp} - \text{proj}_{ZML}(p_{csp}) \|_2
\end{equation}

\noindent where $\mathcal{S}$ denotes the support polygon of current contact points,  and \( \text{proj}_{ZML}(p_{csp}) \) is the projected ZML point in the horizontal plane.  If $d \in [0, 0.32]$, the ZML point is inside the support polygon; if $d \in (0.32, 0.36]$, it lies near the edge of support polygon; and if $d \in (0.36, +\inf)$, the ZML has exited the support region, indicating potential instability. It is worth noting that the closer $d$ is to 0, the system is more stable and satisfy the ZMP constraint.

\begin{figure}[htbp]
\centering
\vspace{-5pt}
\includegraphics[width=0.95\linewidth]{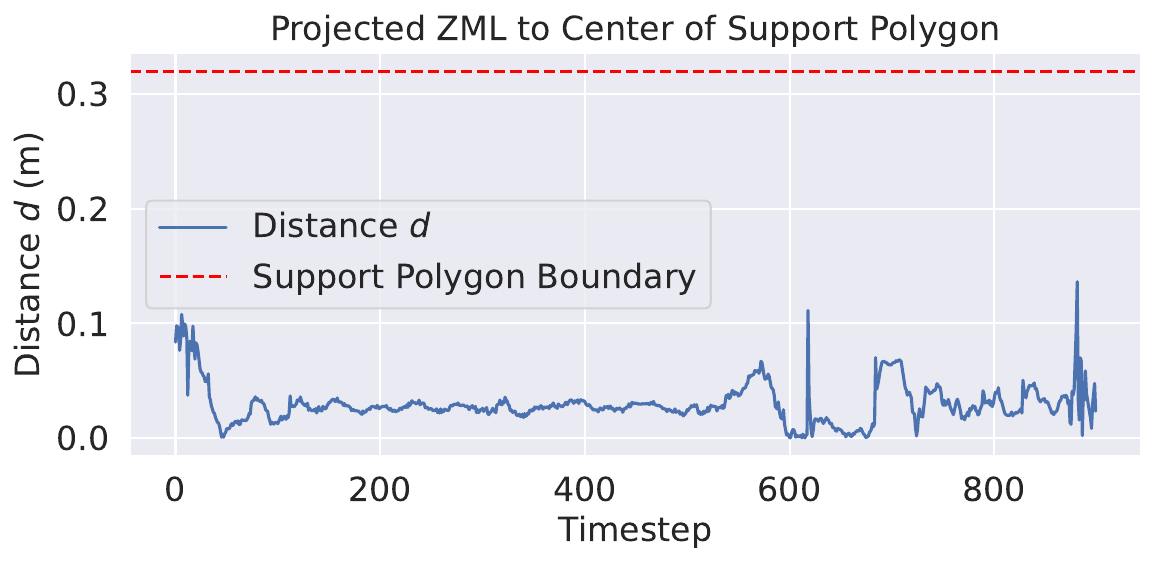}
\caption{\textbf{Stability metric $d$ over time.} The curve shows the distance between the ZML projection and the center of support polygon. The distance remains $\le 0.32$, confirming that the ZMP condition is always satisfied.}
\label{fig::d_curve}
\vspace{-10pt}
\end{figure}

To verify that our TOP satisfies this condition, \cref{fig::d_curve} shows the time series of $d$ for representative tasks. The results confirm that $d$ consistently remains below 0.32, proving that the ZML stays well inside the support polygon during dynamic execution, even with rapid upper-body motions.

These results theoretically validate that our closed-loop controller robustly maintains dynamic stability by keeping the ZMP within feasible bounds, complementing the empirical results and reinforcing the soundness of our framework.

\subsection{Analysis of Momentum}
To rigorously analyze how fast upper-body motions can induce momentum disturbances and compromise whole-body stability, we provide both theoretical and empirical evidence. 

First, the total angular momentum $\bm{L}$ of the robot can be expressed as the sum of the floating base angular momentum and the contribution from each upper-body joint:
\begin{equation}
\bm{L} = \bm{I}_B \bm{\omega}_B + \sum_{i=1}^{n} I_i \dot{q}_i,
\end{equation}
where $\bm{I}_B$ and $\bm{\omega}_B$ denote the inertia tensor and angular velocity of the base, respectively, and $I_i$, $\dot{q}_i$ are the inertia and velocity of joint $i$. Taking the time derivative shows that rapid changes in joint accelerations $\ddot{q}_i$ directly induce compensatory reactions on the base:
\begin{equation}
\bm{I}_B \dot{\bm{\omega}}_B = -\sum_{i=1}^{n} I_i \ddot{q}_i.
\end{equation}

This relationship indicates that large joint accelerations generate equal and opposite angular impulses on the floating base, which can destabilize the robot if left unregulated. Integrating over time confirms that the base’s angular drift is proportional to the accumulated joint accelerations:
\begin{equation}
\Delta \bm{\omega}_B \propto -\int_{t_0}^{t} \sum_{i=1}^{n} I_i \ddot{q}_i \, dt.
\end{equation}

Moreover, faster joint velocities $\dot{q}_i$ also amplify the total angular momentum, further stressing the need for regulating motion speed. To address this, our Time Optimization Policy (TOP) explicitly shapes the motion timing to keep both $\dot{q}_i$ and $\ddot{q}_i$ within safe bounds, thereby suppressing sudden momentum spikes and preventing undesired base rotations.

\begin{figure}[htbp]
    \centering
    \vspace{-5pt}
    \includegraphics[width=1.0\linewidth]{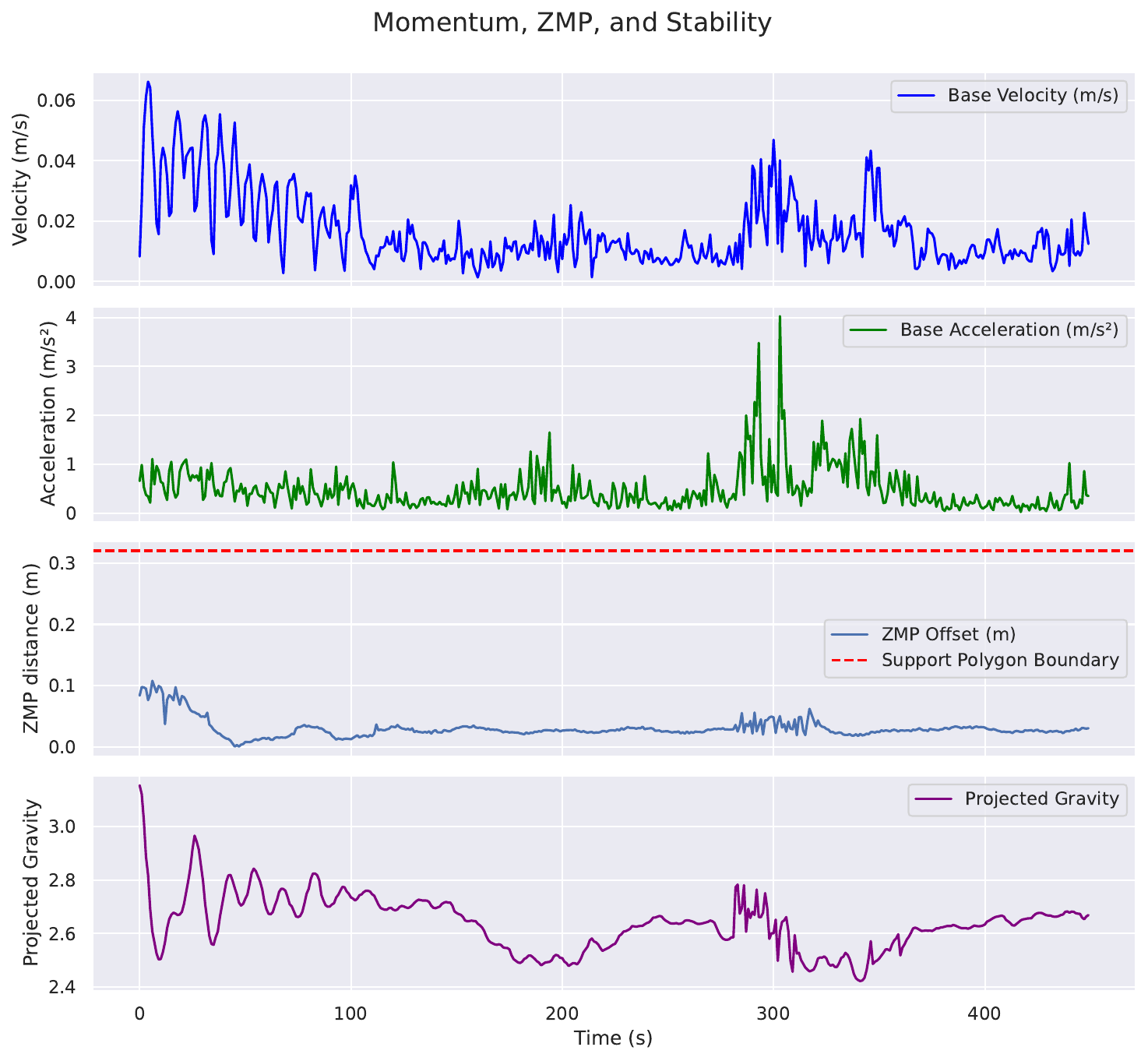}
    \caption{\textbf{Momentum, ZMP and stability.} The figure shows the measured floating base velocity and acceleration during a fast upper-body motion. The TOP smooths out large momentum variations, helping to keep the ZMP within the support region and maintain balance.}
    \label{fig::momentum}
    \vspace{-10pt}
\end{figure}

To validate this mechanism in practice, \cref{fig::momentum} shows the measured floating base velocity and acceleration during a fast upper-body motion. The results demonstrate that the TOP effectively smooths out abrupt base movements by adaptively slowing the arm when the momentum is excessive. This mitigation helps maintain the Zero Moment Point (ZMP) inside the support polygon, ensuring stable whole-body balance.

Overall, this momentum analysis clarifies why simply decoupling the upper-body motion without proper time optimization can lead to significant base disturbance, and highlights the necessity of regulating joint accelerations and velocities to maintain dynamic stability.

\subsection{Training Efficiency}

To provide a fair assessment of the training cost of our approach, we report the training time of each module and compare the overall pipeline with relevant baselines such as Exbody, OmniH2O, and Mobile-Television. 

Specifically, our framework requires approximately 10 hours to train the VAE motion prior, and about 5 hours each for the reinforcement learning (RL) policy and the time-optimized policy (TOP) module, when trained on a single NVIDIA RTX 4090 GPU. In total, our pipeline completes training in roughly 20 hours.

By contrast, training Exbody and OmniH2O under the same hardware setting takes around 22–32 hours, as these baselines require longer joint optimization of imitation and policy modules without task-specific time adaptation. As summarized in \cref{tab:training_time}, our method achieves comparable or better training efficiency while providing adaptive timing and stable manipulation capabilities.

\begin{table}[htbp]
\centering
\caption{Comparison of Training Efficiency}
\label{tab:training_time}
\vspace{-5pt}
\resizebox{0.47\textwidth}{!}{
\begin{tabular}{lcc}
\toprule
\textbf{Method} & \textbf{Hardware} & \textbf{Total Training Time (hrs)} \\
\midrule
\textbf{VAE} & 1x RTX 4090 (24GB) & 10 (approx.) \\
\textbf{RL}  & 1x RTX 4090 (24GB) & 5 (approx.) \\
\textbf{TOP} & 1x RTX 4090 (24GB) & 5 (approx.) \\
\textbf{Ours (VAE + RL + TOP)} & 1x RTX 4090 (24GB) & 20 (approx.) \\
Exbody\cite{cheng2024expressive} & 1x RTX 4090 (24GB) & 25 (approx.) \\
OmniH2O\cite{he2024omnih2o} & 1x RTX 4090 (24GB) & 32 (approx.) \\
Mobile-Television\cite{lu2024mobile} & 1x RTX 4090 (24GB) & 22 (approx.) \\
\bottomrule
\end{tabular}
}
\vspace{-10pt}
\end{table}

\subsection{Analysis of Lower-Body Policy Changes}

To clarify the dependency of the Time Optimization Policy (TOP) on the pretrained lower-body controller, we conducted ablation experiments by varying the lower-body policy in two ways: (1) training the lower-body RL policy with a reduced dataset $\mathcal{M'}$ that contains only 50\% of the motion clips, and (2) swapping in a lower-body policy pretrained with a different encoder structure $E'$, which reduces the VAE window size from 30 to 15 while keeping the latent dimension unchanged at 64 to maintain observation consistency.

\begin{table}[htbp]
\centering
\vspace{-5pt}
\caption{Robustness of the TOP policy under different lower-body controllers and encoders.}
\label{tab:top_dependency}
\resizebox{0.48\textwidth}{!}{
\begin{tabular}{l|c|c|c|c}
\toprule
\textbf{Methods} & \textbf{Success Rate (\%)} $\uparrow$ & $\mathbf{E}^{upper}_{jpe}$ $\downarrow$ & $\mathbf{E_{g}}$ $\downarrow$ & $\mathbf{E}^{upper}_{acc}$ $\downarrow$ \\
\midrule
Baseline & 95.30 & 0.0269 & 2.729 & 10.41 \\
RL with dataset $\mathcal{M'}$ & 89.40 & 0.0338 & 3.545 & 10.94 \\
RL with encoder $E'$ & 91.85 & 0.0301 & 3.312 & 10.65 \\
\bottomrule
\end{tabular}
}
\vspace{-10pt}
\end{table}

These results show that while the TOP can partially adapt to moderate changes in the lower-body dynamics, its performance degrades if the lower-body policy is trained with a reduced dataset or uses a different encoder. Specifically, the success rate drops by approximately 4–6\% and the upper-body joint tracking errors increase, indicating that the synergy between the TOP and the lower-body controller benefits from a shared motion prior and consistent encoder design.

This design choice is intentional: by first training a robust lower-body controller with domain randomization and a consistent motion representation, we ensure stable base stabilization under diverse upper-body motions. The TOP then learns to optimize time allocation to best fit the stabilizer’s dynamics. 

Although the framework does not assume rigid coupling, it does rely on compatible motion encodings to maintain robust performance. For future work, we plan to investigate end-to-end co-training or meta-adaptation strategies to further reduce this dependency.

\end{document}